\documentclass[lettersize,journal]{IEEEtran}
\usepackage{subfigure}
\usepackage{amsmath,amsfonts}
\usepackage{algorithmic}
\usepackage{algorithm}
\usepackage{array}
\usepackage[caption=false,font=normalsize,labelfont=sf,textfont=sf]{subfig}
\usepackage{textcomp}
\usepackage{stfloats}
\usepackage{url}
\usepackage{verbatim}
\usepackage{graphicx}
\usepackage{cite}
\usepackage{threeparttable}
\usepackage{array}
\usepackage{multirow}

\newtheorem{assumption}{Assumption}

\begin{document}

\title{Domain Adversarial Active Learning for Domain Generalization Classification}

\author{Jianting Chen, Ling Ding, Yunxiao Yang, Zaiyuan Di, and Yang Xiang
\thanks{Jianting Chen, Ling Ding, Yunxiao Yang, Zaiyuan Di, and Yang Xiang are with the college of Electronic and Information Engineering, Tongji University, Shanghai 201804, China. Email:  \{tj\_chenjt, dling, yang\_98, 2331896\}@tongji.edu.cn, tjdxxiangyang@gmail.com. 
}}

\markboth{Journal of \LaTeX\ Class Files,~Vol.~14, No.~8, August~2021}%
{Shell \MakeLowercase{\textit{et al.}}: A Sample Article Using IEEEtran.cls for IEEE Journals}


\maketitle

\begin{abstract}
Domain generalization models aim to learn cross-domain knowledge from source domain data, to improve performance on unknown target domains.
Recent research has demonstrated that diverse and rich source domain samples  can enhance domain generalization capability.
This paper argues that the impact of each sample on the model's generalization ability varies.
Despite its small scale, a high-quality dataset can still attain a certain level of generalization ability.
Motivated by this, we propose a domain-adversarial active learning (DAAL) algorithm for classification tasks in domain generalization.
First, we analyze that the objective of tasks is to maximize the inter-class distance within the same domain and minimize the intra-class distance across different domains.
To achieve this objective, we design a domain adversarial selection method that prioritizes challenging samples. 
Second, we posit that even in a converged model, there are subsets of features that lack discriminatory power within each domain.
We attempt to identify these feature subsets and optimize them by a constraint loss.
We validate and analyze our DAAL algorithm on multiple domain generalization datasets, comparing it with various domain generalization algorithms and active learning algorithms.
Our results demonstrate that the DAAL algorithm can achieve strong generalization ability with fewer data resources, thereby reducing data annotation costs in domain generalization tasks.
\end{abstract}


\section{Introduction}

\IEEEPARstart{T}{he} deep domain generalization task is an extension of deep learning beyond the i.i.d. assumption. 
It falls within the research domain of domain transfer learning. 
Conventional domain transfer learning algorithms are applied in scenarios involving multiple domain data. 
The purpose of domain transfer learning algorithms is to transfer knowledge from a set of source domain data to improve generalization ability on the target domain. 
Domain generalization, a more advanced version of domain transfer learning, aims to establish a  model based on other related source domains, which is independent of data from target domains. 
Specifically, in domain generalization tasks, the training and test sets consist of non-overlapping domains. 
It is expected that the model trained on the training set performs as well as possible on the test set.
Domain generalization models can handle unknown data distributions and are suitable for dynamic environments. 
Thus, these models have significant practical value in fields like healthcare and industry. 

Our research focuses on deep learning classification models in the context of domain generalization. 
We argue that the main issue stems from the unpredictable outputs of models when confronted with unknown domains.
This unpredictability arises from the divergence in test data distributions, resulting in a covariate shift in the features.
The classifier faces a challenge in accurately identifying these shifted features.
In addressing this challenge, the primary objective of classification models is to minimize the intra-class distance while maximizing the inter-class distance.
By minimizing the intra-class distance of features across different domains, models output domain-insensitive feature representations,  thereby alleviating the shift in features from unknown domains.
Maximizing the inter-class distance allows models to expand the feature range of each category, enhancing the classifier's tolerance and robustness towards shifted features. 


Current research on deep domain generalization can be divided into three categories \cite{DBLP:journals/tkde/WangLLOQLCZY23}.
The first category is data augmentation. 
Style transfer \cite{DBLP:conf/iclr/ZhouY0X21}, mixup \cite{DBLP:conf/eccv/ManciniARC20}, and adversarial generation \cite{DBLP:conf/aaai/ZhouYHX20} techniques are used to generate diverse samples.
Within more domain samples, models can generate more distinctive features, leading to robust decision boundaries.
Data augmentation is particularly useful when the training set consists of only a single domain, as it can simulate new domain samples \cite{DBLP:conf/cvpr/ChoiDCYPY23,DBLP:conf/cvpr/LiGCHWMYLX21,DBLP:conf/cvpr/QiaoZP20}.
However, Data augmentation methods are constrained by the type of inputs, and complex data augmentation algorithms might require additional training overhead.
The second category focuses on representation learning.
These methods aim to disentangle or align features to learn domain-invariant representations. 
Features that remain invariant despite shifts in the source domain should also be resilient to shifts in any unseen target domain \cite{DBLP:conf/nips/Ben-DavidBCP06}. 
While these methods do exhibit a certain degree of generality, they require high-quality and diverse data in the source domain.
The third category focuses on learning strategies, which mainly includes ensemble learning \cite{DBLP:conf/dagm/DInnocenteC18}, meta learning \cite{DBLP:conf/cvpr/0009GWL23}, gradient operations \cite{DBLP:conf/eccv/HuangWXH20,DBLP:conf/iccv/MansillaEMF21}, and self supervision \cite{DBLP:conf/cvpr/CarlucciDBCT19,DBLP:conf/iccv/KimYPKL21}.
These methods enhance the learning process and render models insensitive to domains.
However, the complexity presents a challenge to the effectiveness of these techniques.

\begin{figure}[!t]
  \centering
  \subfigure[Artpainting]{
   \includegraphics[width=0.45\linewidth]{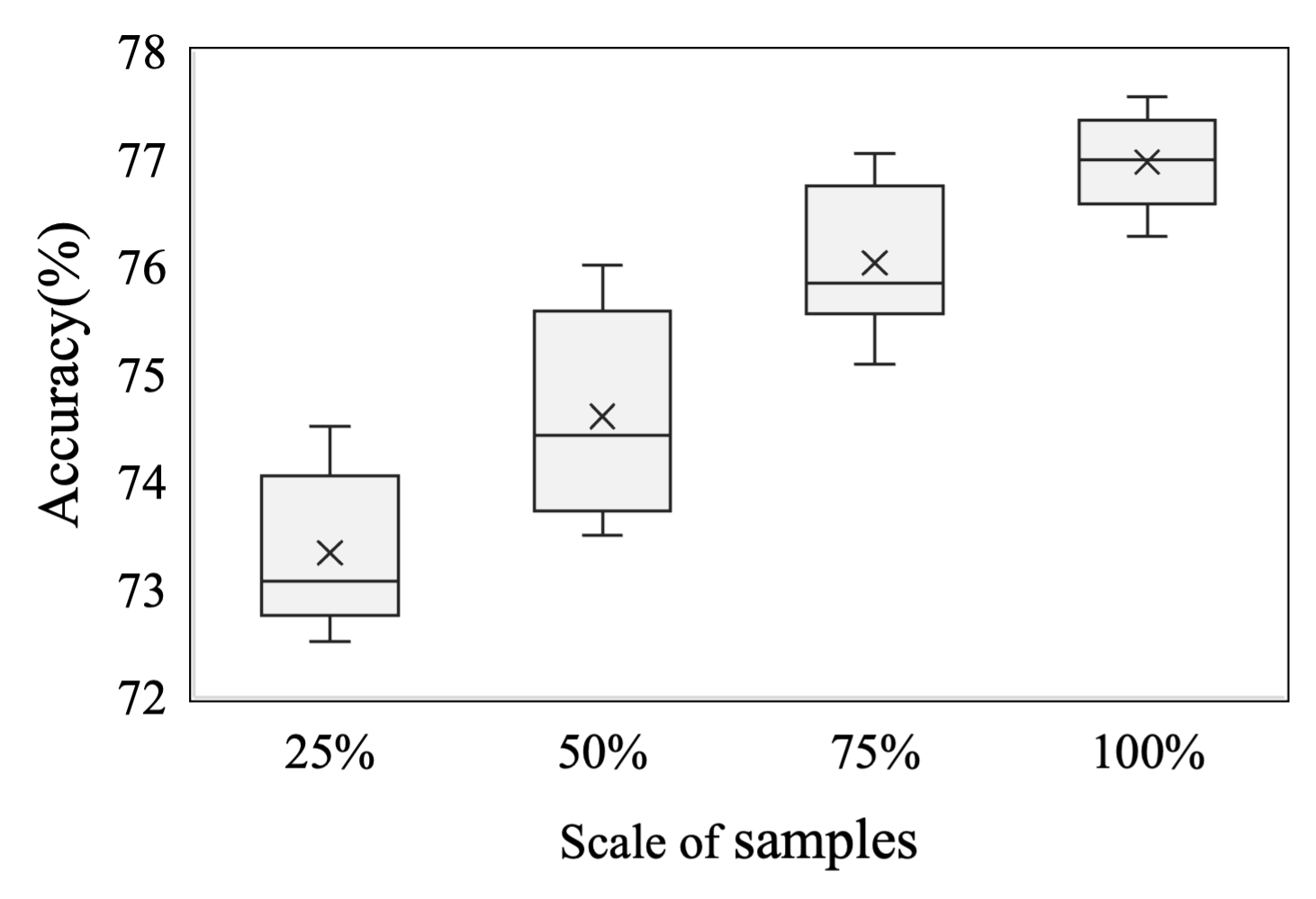}
  }
 \subfigure[Cartoon]{
   \includegraphics[width=0.45\linewidth]{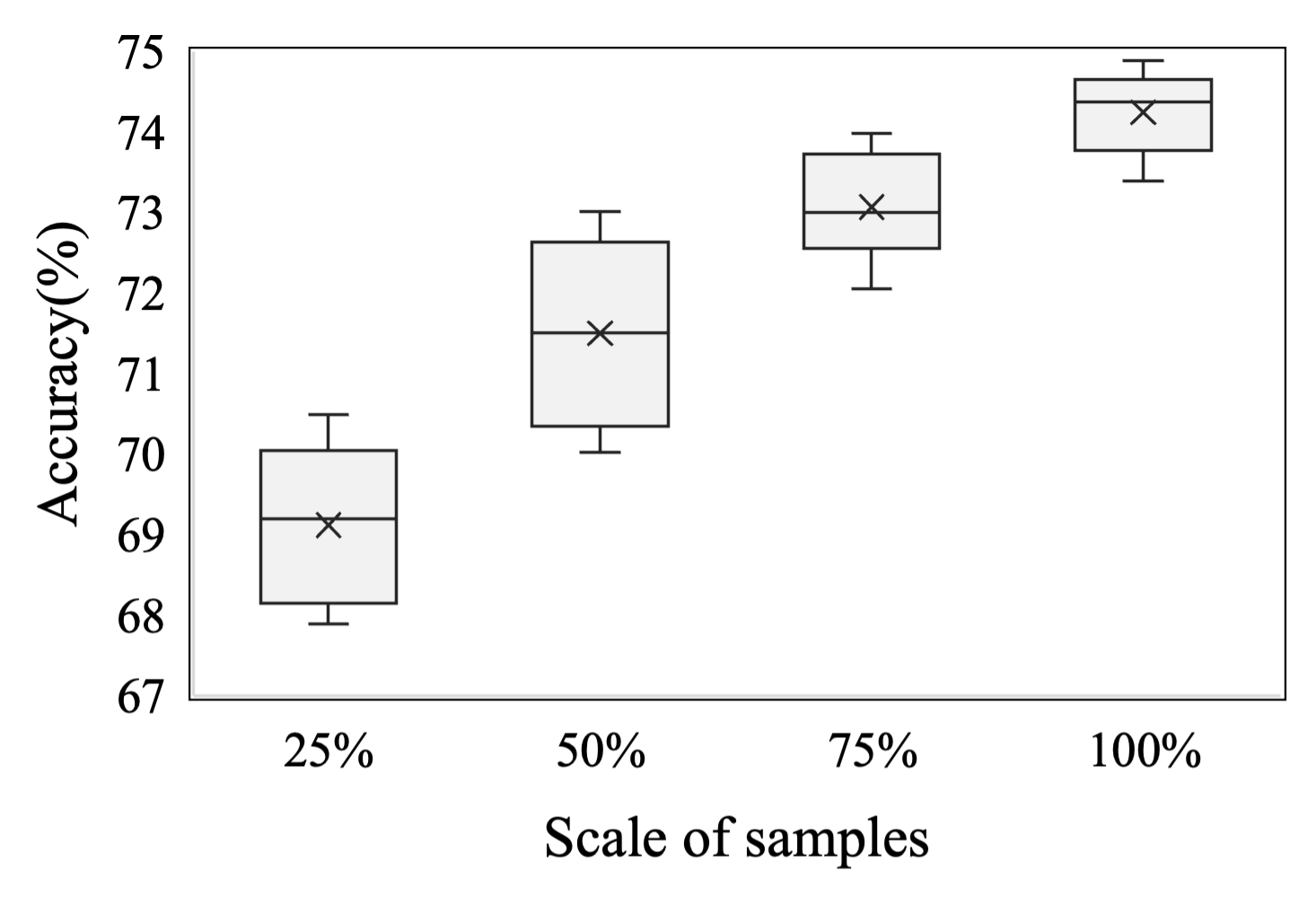}
  }
  \caption{
The domain generalization experiment results on the PACS dataset were obtained by randomly selecting training subsets. 
We conducted the experiment using four different sizes of training subsets - 25\%, 50\%, 75\%, and 100\%. 
The ERM models were trained accordingly.
  }
  \label{fig:boxes}
\end{figure}

Recurrent research has demonstrated that diverse samples from multiple domains contribute to enhancing domain generalization capability. 
In this paper, we focus primarily on quality rather than quantity.
We argue that different samples in the training set play diverse roles in enhancing the generalization. 
The result of the validation experiments (see Fig. \ref{fig:boxes}) substantiates our viewpoint. 
Randomly selected samples were used to create subsets with different sizes, which were then utilized to train the model.
During repeated experiments, small training subsets resulted in a higher standard deviation in test results, due to the variations in the samples included. 
Occasionally, test results from small training subsets even outperformed those from larger training subsets.
This indicates that there are some training subsets  that can possess greater generalization ability. 
Our objective is to employ algorithms to select cost-effective training samples, thus enhancing models and minimizing data expenses.

For this objective, we aim to integrate domain generalization tasks with active learning. 
Active learning is a type of semi-supervised algorithm that actively selects the most informative samples for labeling to enhance model performance. 
The active learning algorithm involves three fundamental steps: sample selection, expert labeling, and model training, which are iteratively executed to improve models. 
To tailor it to domain generalization tasks, we will optimize the steps of sample selection and model training. 

Specifically, we develop a domain adversarial active learning  (DAAL)  algorithm. 
Regarding sample selection, we devise a domain adversarial selection method. 
The selected adversarial samples aim to minimize the intra-class distance within the same domain and maximize the inter-class distance across different domains, which opposes the objective of models.
The underlying expectation is that models can handle these challenging samples, thereby improving generalization abilities. 
Regarding model training,  we propose an method to maximize the inter-class distance by optimizing the features with weak discrimination. 
At the start of each iteration,  we identify subsets of features within each domain that exhibit insufficient inter-class distances. 
Then, we incorporate an additional optimization loss to enhance the discriminatory ability of these feature subsets.
By employing these two methods, the model can effectively enhance its generalization abilities. 
To validate our DAAL algorithm, we conducted experiments on multiple domain-generalization datasets. 
In comparison with other domain generalization algorithms, the DAAL algorithm yielded comparable results while requiring less data. 
Furthermore, the DAAL algorithm outperformed other active learning algorithms in terms of generalization abilities. 
Lastly, through T-SNE visualization, we observed a significant improvement in intra-domain class boundaries, affirming the effectiveness of our methods.

The main contributions of our work can be summarized as follows:

\begin{enumerate}
\item [1] We integrate domain generalization tasks with an active learning framework.
We design a domain adversarial sample selection method that can identify challenging samples that is benefit to the model's generalization ability.
\item [2] We optimize the loss function based on the iterative training process.
We introduce a method to actively identify and optimize feature subsets with insufficient inter-domain distances, enhancing robustness across different domains.
\item [3] We conduct experiments on various domain generalization datasets and demonstrate that our approach achieves higher sample efficiency compared to other domain generalization algorithms.
\end{enumerate}

\section{Related work}
\subsection{Domain Generalization}

Many studies have been conducted on domain generalization tasks, exploring various techniques including data augmentation, representation learning, and learning strategies.

Data augmentation is a  widely used technique in deep learning to enhance model generalization. 
Random augment is a simple approach, whereas RandConv \cite{DBLP:conf/iclr/XuLYRN21} and Pro-RandConv \cite{DBLP:conf/cvpr/ChoiDCYPY23} leverage a random parameterized convolutional layer to transform images into novel domains. 
Inspired by adversarial gradients, Adaptive Data Augmentation (ADA) \cite{DBLP:conf/nips/VolpiNSDMS18,DBLP:conf/cvpr/QiaoZP20} and CrossGrad \cite{DBLP:conf/iclr/ShankarPCCJS18} perturb input data using gradients acquired from label classifiers and domain classifiers.
Style transfer models such as AdaIN \cite{DBLP:conf/iccv/HuangB17} can be employed to transfer data between diverse domains \cite{DBLP:conf/iccv/YueZZSKG19}. 
Generative models like Generative adversarial networks (GAN) and variational auto-encoders (VAE) have also been used to create synthetic inputs \cite{DBLP:conf/wacv/RahmanFBS19,DBLP:conf/cvpr/QiaoZP20}.
The Progressive Domain Expansion Network (PDEN) \cite{DBLP:conf/cvpr/LiGCHWMYLX21} employs subnetworks to progressively generate simulated domains for single domain generalization.
Mixup \cite{DBLP:conf/iclr/ZhangCDL18}, a well-established technique for image augmentation, has gained popularity in domain generalization tasks \cite{DBLP:conf/cvpr/ShuCW0L21}.
Mixup can also be employed to mix features from various domains, proving to be effective in enhancing generalization capabilities \cite{DBLP:conf/iclr/ZhouY0X21,DBLP:conf/cvpr/XuZ0W021,DBLP:conf/cvpr/Qiao021}.
The primary challenge in data augmentation is finding the right balance between generating diverse samples while maintaining their quality.
Hence, complex data augmentation methods require careful control over this balance.

Domain-invariant representation learning is widely practiced to address domain generalization. 
By explicitly aligning feature distributions, it is possible to achieve consistent feature representations across diverse domains.
Maximum Mean Discrepancy (MMD) \cite{DBLP:conf/cvpr/LiPWK18} and Wasserstein distance \cite{DBLP:journals/ijon/ZhouJSWC21} are used to measure distribution consistency. 
Domain adversarial learning aims to minimize differences in features across domains by training a domain discriminator to render the features indistinguishable \cite{DBLP:conf/eccv/LiTGLLZT18,DBLP:conf/cvpr/GongLCG19}.
Feature disentanglement methods aim to separate label-related and domain-related factors from a set of mutually independent features.
In related research, the causal invariance principle is considered as the main basis for disentangling features.
For example, StableNet \cite{DBLP:conf/cvpr/Zhang0XZ0S21} employs Fourier features as causal factors and  ensures their independence.
Causality Inspired Representation Learning (CIRL) \cite{DBLP:conf/cvpr/LvLLZLWL22} identifies causal factors using a causal intervention module. 
Invariant Risk Minimization (IRM) \cite{DBLP:journals/corr/abs-1907-02893} optimizes the model's loss to achieve consistent classification across domains.
A range of theoretical analyses and variant methods have been proposed to further optimize IRM \cite{DBLP:conf/iclr/RosenfeldRR21,DBLP:conf/icml/LiuH00S21,DBLP:conf/cvpr/LinDWZ22}.
The findings obtained from these methods indicate that domain-invariant features are crucial for effective domain generalization.
However, excessively strong invariance constraints can impact the learning of label semantics.
It is important to balance the strength of invariance constraints and label semantics.

Various learning strategies have been employed to enhance the domain generalization of models.
The study conducted by \cite{DBLP:conf/icip/ManciniBC018}  involved multiple domain-specific classifiers and aggregate their predictions to boost performance.
Meta-Learning for Domain Generalization (MLDG)\cite{DBLP:conf/aaai/LiYSH18} attempts to learn a general model by dividing multiple source domains into meta-train and meta-test sets to simulate domain shift.
Self-supervised learning methods, with contrastive learning as a representative, are also widely applied in domain generalization \cite{DBLP:conf/mm/JeonHLLB21}.
For instance, SelfReg \cite{DBLP:conf/iccv/KimYPKL21} utilizes positive sample pairs to construct contrastive loss and  effectively avoids representation collapse.
Gradient information can provide guidance for optimizing domain generalization.
Representation Self-Challenging (RSC) \cite{DBLP:conf/eccv/HuangWXH20} utilizes gradients to identify salient features and encourages the remaining features to represent label semantics. 
Neuron Coverage-guided Domain Generalization (NCDG) \cite{DBLP:journals/pami/TianLX0023} introduces gradient similarity regularization to maximize the neuron coverage of networks. 
Sharpness-Aware Gradient Matching (SAGM) \cite{DBLP:conf/cvpr/WangZLZ23} proposes using empirical risk and perturbed loss as learning objectives and aligning the gradient directions of both objectives.
Additionally, optimization methods for learning strategies often complement data augmentation and representation learning methods, enabling better utilization of training data from multiple domains.
The work presented in this paper falls within this category.

\subsection{Active Learning}

Deep active learning algorithms are primarily categorized into two groups: uncertainty-based algorithms and diversity-based algorithms, depending on the sampling strategy used. 
Additionally, hybrid techniques that leverage the strengths of both methods are available \cite{DBLP:journals/csur/RenXCHLGCW22}.

Uncertainty-based active learning is a simple and computationally efficient algorithm. 
After the initial training of the deep learning model, this algorithm identifies high-uncertainty samples and selects them to train the model. 
The focus of uncertainty-based algorithms is the measurement of sample uncertainty. 
The least confidence \cite{DBLP:conf/ijcnn/WangS14} algorithm selects the sample that has the lowest posterior probability for its most likely label.  
A lower probability corresponds to a higher uncertainty. 
The maximum entropy \cite{2010Active}  algorithm calculates the information entropy based on the posterior probability and selects the sample with highest entropy value. 
Bayesian active learning by disagreement (BALD) \cite{DBLP:conf/icml/GalIG17}  uses dropout to compute multiple posterior probabilities for each sample and estimates the mutual information as a guide for sample selection. 
Batch active learning by diverse gradient embeddings (BADGE) \cite{DBLP:conf/iclr/AshZK0A20}  selects samples with higher gradients on the classifier parameters to form the subsequent training set.	

Diversity-based active learning leverages the concept of a core set, which aims to identify a subset of samples that capture the distribution pattern of the entire dataset. 
Common methods employed in diversity-based active learning include clustering algorithms \cite{DBLP:conf/nips/CitovskyDGKRRK21}, exploration-exploitation algorithms \cite{DBLP:conf/icdm/YinQCLWZD17}, and core set algorithms \cite{DBLP:conf/iclr/SenerS18}. 
Discriminative models \cite{DBLP:journals/kbs/ZhouSYHWC21} can also be employed in the diversity-based algorithm. 
The algorithm selects representative unlabeled samples that the model cannot distinguish. 
However, due to the complex distribution patterns in high-dimensional input data in deep learning tasks, choosing representative samples directly can be challenging. 
As a result, diversity-based algorithms tend to be more complex and computationally intensive than uncertainty-based algorithms.

To leverage the benefits of above methods, they can be simultaneously applied to the sample selection process. 
For instance, the BADGE \cite{DBLP:conf/iclr/AshZK0A20} algorithm adopts a two-stage strategy that considers both diversity and uncertainty in sample selection. 
Weighted averaging can also be used to integrate the two methods in some studies \cite{DBLP:conf/icdm/YinQCLWZD17}. 

\section{Problem description}
\label{problem_description}
In this section, we will introduce the problem description and notations.
A domain generalization task divides the data into different domains, represented by $e\in E$. 
These domains have distributional differences, meaning that the joint distribution of input samples $X$ and output labels $Y$ is different across domains, i.e., $P(X^iY^i) \neq P(X^jY^j)$ for $i \neq j$.
The training samples are from the domain set $E_{tr}$, while the test samples are from other unknown domain set denoted as $E_{te}$.

In the active learning scenario, all available unlabeled data samples are stored in a data pool, denoted as $U_{tr} = \{X^e | e \in E_{tr}\}$.
The active learning process consists of $\tau$ rounds of learning opportunities.
Each round involves selecting $n$ samples from the data pool $U$ and obtaining their label annotations from experts, resulting in labeled samples $B_{tr}^{(t)} = \{X^{(t)}, Y^{(t)}\}$.
The model is then trained based on all the labeled samples collected so far, denoted as $D_{tr} = \bigcup_{i}^{t}B_{tr}^{(i)}$. 
The objective is to improve learning efficiency by selecting appropriate samples for labeling and training, thereby reducing resource wastage on invalid samples.

In this paper, we focus on multi-class classification tasks using deep learning models..
The set of labels is denoted as $K$.
Each labeled sample consists of an input $x_i$ and its corresponding label $y_i$.
The deep learning model processes the input $x_i$ and produces an output $z_i = h(x_i)$.
The goal of our research is to enhance the model's ability to generalize across domains by selecting valuable samples through active learning.
This involves two objectives: minimizing the error on the unknown domain test set, denoted as $\min{\mathbb{E}}_{(x_i, y_i)\sim D_{te}}[\mathcal{L}(z_i, y_i)]$, and minimizing the number of labeled samples, denoted as $\min{(|D_{tr}|)}$.

To facilitate subsequent description, we decompose the deep learning model into a feature extractor $f_i=f(x_i)$ and a classifier $z_i=g(f_i)$. 
The feature extractor transforms a input $x_i$ into a feature representation $f_i \in \mathbb{R}^d$, where $d$ is the dimensionality of the feature vector.
Each dimension in the feature vector is denoted as $f_i^l$, representing the feature value in the $l$-th dimension. 
The set of indices for all the features is denoted as $F=\{1,2,\ldots,d\}$.

\begin{figure}[!t]
  \centering
   \includegraphics[width=1.0\linewidth]{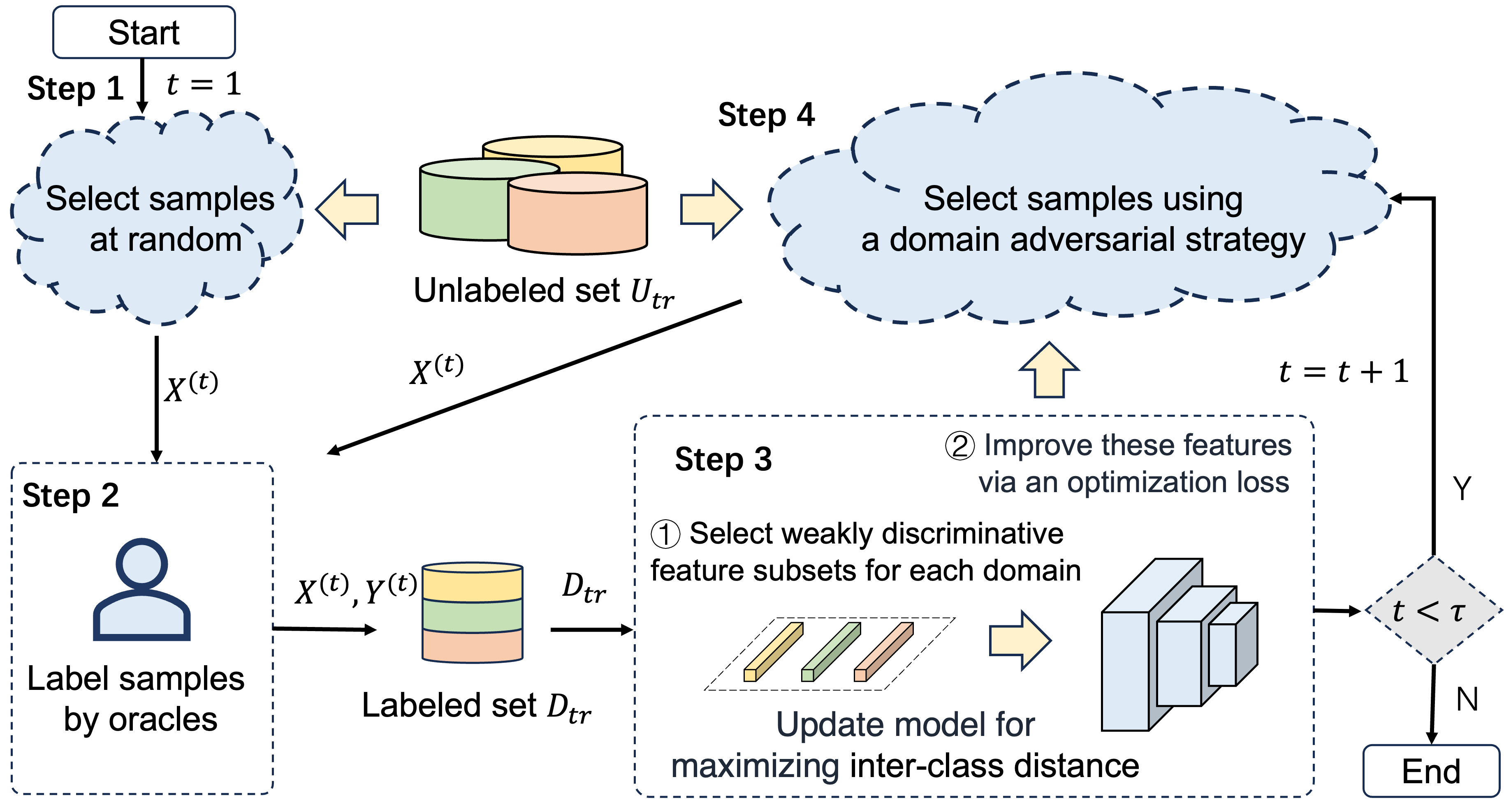}
  \caption{
The iterative framework of the domain adversarial active learning algorithm.
  }
  \label{fig:framework}
\end{figure}

\section{Method}
In this section, we will describe the motivation and concrete implementation of our DAAL algorithm. 
The main focus is to improve domain generalization in a active learning scenario.
The framework of our DAAL algorithm is illustrated in Fig. \ref{fig:framework}.
The following steps are involved in our method:

\textbf{Step 1: Initialization phase.}
In this step, since there is no reference available, we randomly select $n$ samples, denoted as $X^{(1)}$, from the sample pool $U_{tr}$. 

\textbf{Step 2: Expert labeling.}
The samples selected in Step 1 and Step 4 are labeled with categories by experts. 
This forms a labeled set, denoted as $B_{tr}^{(t)}=\{X^{(t)},Y^{(t)}\}$. 
The newly labeled samples are then added to the training set $D_{tr}$.

\textbf{Step 3: Model training.}
This step involves training the model using the updated training set. 
In our DAAL algorithm, we design an maximizing inter-class distance method specifically for domain generalization. 
(See Section \ref{maximizing_inter-class_distance_within_the_same_domain}.)

\textbf{Step 4: Sample selection.}
We select $n$ samples, denoted as $X^{(t+1)}$, from $U_{tr}$ based on the trained model. 
For domain generalization, we design a domain adversarial selection method to prioritize challenging samples. 
(See Section \ref{domain_adversarial_sample_selection}.) If the iteration continues, set $t=t+1$ and proceed to Step 2.

By optimizing the sample selection and model training steps, we improve the model's domain generalization ability under limited labeling resources.

\subsection{Domain Adversarial Sample Selection}
\label{domain_adversarial_sample_selection}
The objective of feature learning in multi-classification tasks is to minimize the intra-class distance $\phi_{intra}$ and maximize the inter-class distance $\phi_{inter}$.
The feature distance that satisfies $\phi_{intra}\ll\phi_{inter}$ indicates that the model has formed a reliable decision boundary.
The main idea behind the uncertainty-based active learning is to prioritize the selection of challenging samples.
When the training model can accurately recognize difficult samples, it can also accurately recognize simpler samples.
The model demonstrates generalization ability.

Challenging samples primarily refer to those that the current model cannot recognize.
From the perspective of feature distance, the model's decision boundary on these challenging samples becomes ambiguous.
The inter-class distance decreases, while the intra-class distance increases.
Therefore, the objective of selecting challenging samples can be represented as:

\begin{equation}
\label{eq1}
{\underset {X^{(t)}}{\mathop{argmax}}}{\left[\phi_{intra}(X^{\left(t\right)},Y^{\left(t\right)})-\phi_{inter}(X^{(t)},Y^{(t)})\right]}
\end{equation}
The objective of sample selection contradicts the expected behavior of the learning model, as it seeks to maximize the intra-class distance while minimizing the inter-class distance. 
However, since the sample labels are unknown, the active learning methods prefers to select samples near the decision boundary as challenging samples.

The samples in the domain generalization task are sourced from multiple domains. 
Samples from the same domain have multiple categories, and samples from the same category belong to different domains.
Thus, the intra-class distance is classified as $\phi_{intra}^{same}$ if the samples belong to the same domain, and as $\phi_{intra}^{cross}$ if they belong to different domains.
Similarly, the inter-class distance can also be classified as $\phi_{inter}^{same}$ and $\phi_{inter}^{cross}$.
Furthermore, samples from the same domain exhibit similarity. 
Typically, the distance between features from the same domain is either less than or equal to the distance between features from different domains, denoted as $\phi_{same} \leq \phi_{cross}$.
Thus,  the domain generalization task is expected to have the following relationship regarding feature distance:

\begin{equation}
\label{eq2}
\phi_{intra}^{same}\le\phi_{intra}^{cross}<\phi_{inter}^{same}\le\phi_{inter}^{cross}
\end{equation}

\begin{figure}[!t]
  \centering
   \includegraphics[width=0.6\linewidth]{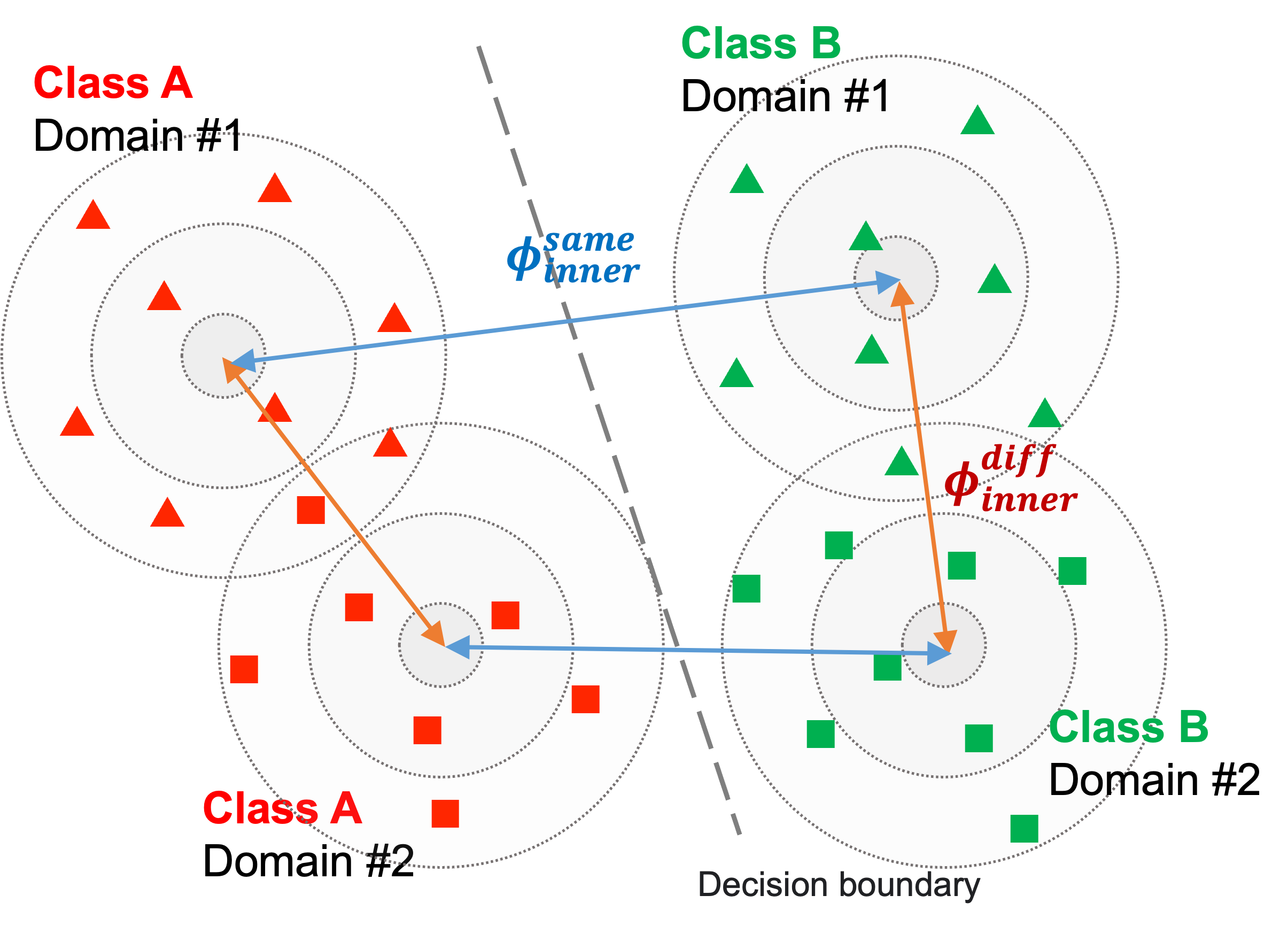}
  \caption{
The illustration of intra-class distance within the same domain and intra-class distance across different domains.
}
  \label{fig:distance}
\end{figure}

An important challenge for this objective is the inequality between $\phi_{intra}^{cross}$ and $\phi_{inter}^{same}$, as illustrated in Fig. \ref{fig:distance}.
If the model features do not satisfy the inequality, 
cross-domain samples may be misclassified.
Hence, the objective of selecting challenging samples in the domain generalization task is defined as

\begin{equation}
\label{eq3}
{\underset {X^{(t)}}{\mathop{argmax}}}{\left[\phi_{intra}^{cross}(X^{\left(t\right)},Y^{\left(t\right)})-\phi_{inter}^{same}(X^{(t)},Y^{(t)})\right]}
\end{equation}
Following this motivation, we propose a sample selection method for domain generalization tasks.

\begin{figure}[!t]
  \centering
   \includegraphics[width=0.6\linewidth]{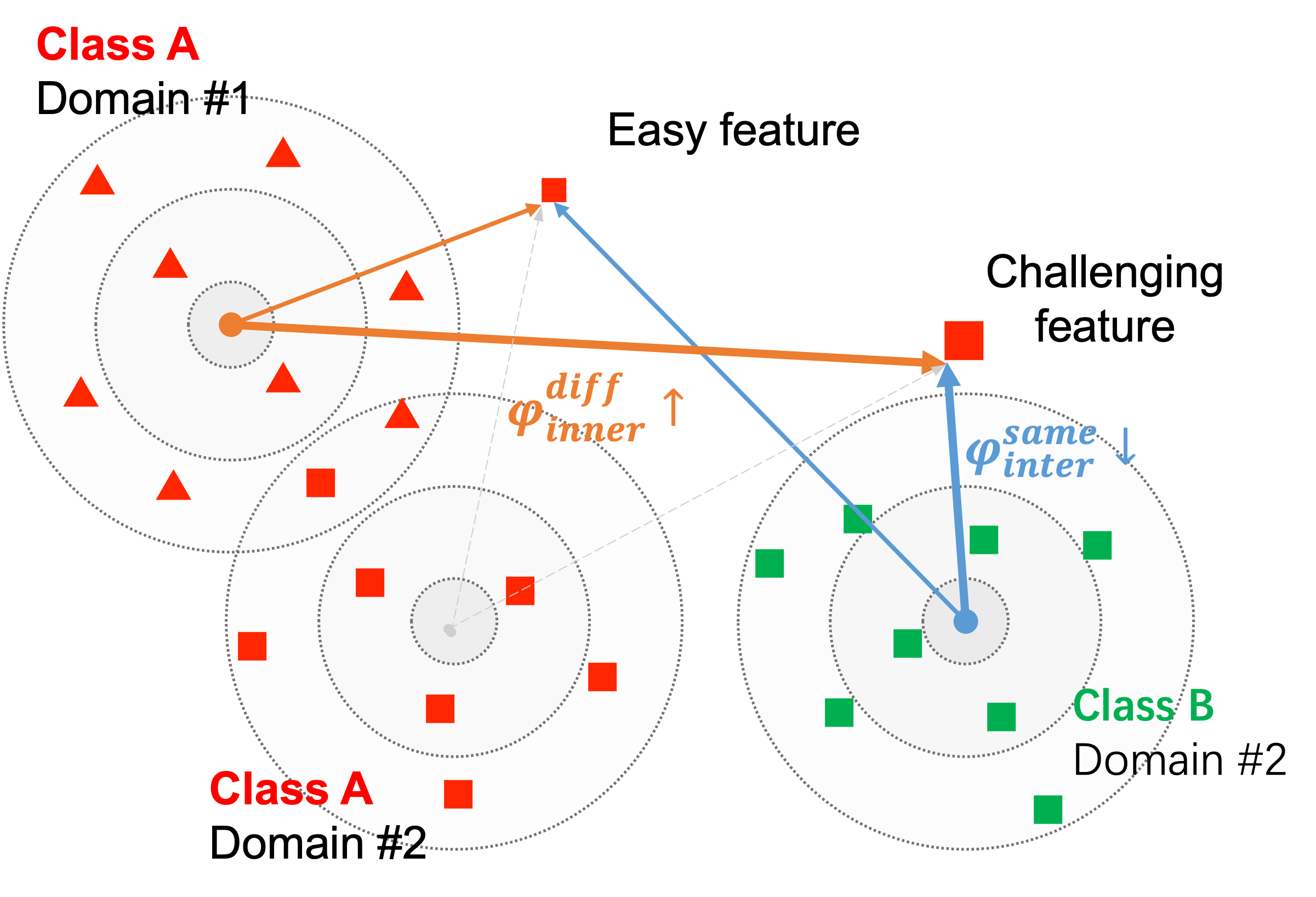}
  \caption{
The illustration of domain adversarial samples.
}
  \label{fig:metric}
\end{figure}

In this method, feature distances serve as the foundation for sample selection. 
However, computing the distance matrix between all pairs of sample features directly incurs a prohibitively high computational cost. 
Hence, we employ an approximate approach. 
Firstly, The centroid of each category within each domain is computed using the labeled data. 
\begin{equation}
\label{eq4}
c_k^a=\frac{1}{n_k^a}\sum_{i}^{|D_{tr}|}{f_i\cdot\mathbb{I}\left(y_i=k,x_i\in X^a\right)}
\end{equation}
Where $n_k^a$ denotes the number of samples in the category  $k$ and from the domain $a$ in the labeled dataset, $\mathbb{I}(\cdot)$ is the indicator function. 
Subsequently, the selection of challenging samples is determined by calculating the distance between the features of unlabeled samples and the domain category centroids, as depicted in Fig. \ref{fig:metric}.

Assume that the sample feature $f_i$ has the label $k$ and is from the domain $a$. 
We calculate the average distance of this feature to the centroid of the same category but from different domains.

\begin{equation}
\label{eq5}
\varphi_{intra}^{cross}(f_i,k,a)=\frac{1}{\left|E_{tr}\right|-1}\sum_b^{|E_{tr}|}{||f_{i},c_{k}^b||\cdot\mathbb{I}}(b\neq a)
\end{equation}
The larger this result is, the greater $\phi_{intra}^{cross}$ after sample selection. 
On the contrary, for $\phi_{inter}^{same}$, we calculate the average distance of this feature to the centroids of different categories but within the same domain.

\begin{equation}
\label{eq6}
\varphi_{inter}^{same}(f_i,k,a)=\frac{1}{\left|K\right|-1}\sum_b^{|K|}{||f_{i},c_{k}^a||\cdot\mathbb{I}}(l\neq k)
\end{equation}
This result exhibits a positive correlation with $\phi_{inter}^{same}$. 
By combining the two, the metric that quantifies the difficulty of the sample is expressed as follows:
\begin{equation}
\label{eq7}
\varphi\left(f_i,k,a\right)=\varphi_{intra}^{cross}(f_i,k,a)-\varphi_{inter}^{same}(f_i,k,a)
\end{equation}

However, since the label $k$ of the sample to be selected is unknown, we can only estimate the probability $p_i$ that the sample belongs to each category  using the existing model. 
Therefore, the measure of sample difficulty is defined as $\varphi(f_i,a) = \mathbb{E}_{k \sim p_i} \varphi(f_i,k,a)$.
We prioritize the $n$ samples with the highest level of difficulty.

In practice, we have observed that combining the domain adversarial selection method with other uncertainty-based selection algorithms yields better results. 
For instance, the initial selection can be performed by choosing $\rho \cdot n$  $(\rho > 1)$ samples from the sample pool using the least confidence algorithm. 
Subsequently, we select $n$ samples from the $\rho n$ candidates using the domain adversarial selection method. 
This combination allows for multiple strategies to be employed simultaneously.

\subsection{Maximizing inter-class distance} 
\label{maximizing_inter-class_distance_within_the_same_domain}
During the training phase, it is expected that the updated model can still satisfy the requirement $\phi_{intra}^{cross} \le \phi_{inter}^{same}$, 
which ensures that the model maintains good generalization performance across different domains.
Given that the inter-class distance and intra-class distance are relative, our objective is to maximize $\phi_{inter}^{same}$. 
This objective enables the model to effectively utilize the feature space.

\begin{figure}[!t]
  \centering
   \includegraphics[width=1.0\linewidth]{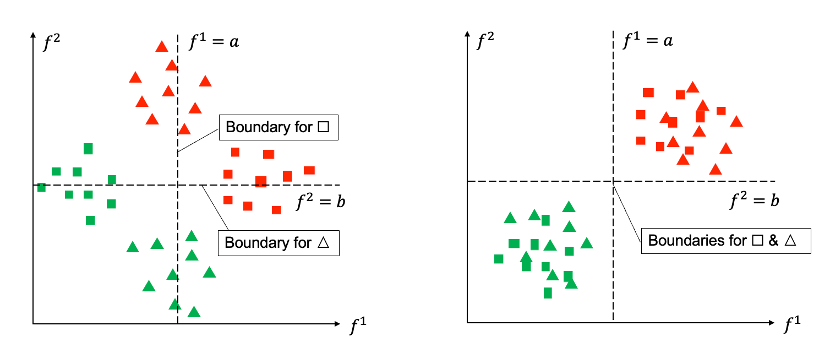}

  \caption{
The potential feature distributions after training with two domain samples.
In the left graph, the square domain establishes a decision boundary based on the feature $f^1$.
The triangle domain establishes a decision boundary with the other feature  $f^2$.
In the right graph, two domains that have the flexibility to rely on either feature for classification.
  }
  \label{fig:boundary}
\end{figure}

Generally, maximizing the inter-class distance can be achieved by using a cross entropy function, which can effectively establish decision boundaries on the training features. 
However, we hypothesize that the decision boundaries may not be reliable and only depend on a subset of features.
This subset of features required for classification varies across different domains, as illustrated in Fig. \ref{fig:boundary}. 
There is a specific subset of features exclusively used for classification in a particular domain. 
The feature subset may not be discriminative in other domains.
For this reason, we propose the following assumption:

\begin{assumption}
\textit{In a given domain $e$, there may still exist a weakly discriminative feature subset $S_e\subset F$ in the output feature set $F$ obtained from the converged model.
Within this feature subset, the features specific to domain $e$ are clustered together, resulting in inadequate inter-class distance.}
\end{assumption}

Based on the assumption, we propose to identify and optimize the weakly discriminative feature subsets for each domain. 

\textbf{Identifying weakly discriminative features}. 
After labeling the new samples and prior to training the model, we aim to identify the weakly discriminative feature subsets for each source domain.
This process results in a collection of feature subsets to be optimized, denoted as $\mathbf{S}=\left\{S_e|e\in E_{tr}\right\}$.

To accomplish this, we propose using a filter feature selection method. 
First, we assess the contribution value $w_y^j$ of each feature $j$ towards the classification of the labeled training set. 
Features with high contribution values are useful across all domains, while those with low contribution values may only be effective in certain domains or have no impact at all. 
The inter-class distance on these low contribution features tends to be insufficient.
Second, we argue that a feature subset exhibiting weak discrimination for a domain $e$ would display a small intra-domain distance. 
To evaluate this, we prepare a binary classification task to identify the sample's domain. 
During this task, we also measure the contribution value $w_e^j$ of each feature $j$. 
In a subset consisting of higher contribution features, the intra-domain distance is smaller.

In terms of implementation, we employ the random forest algorithm to measure the aforementioned contribution values and subsequently normalize the results. 
By taking into account both factors, we calculate a score to determine whether feature $j$ should be considered part of the weakly discriminative feature subset for domain $e$.

\begin{equation}
\label{eq8}
w^j=w_e^j-\alpha w_y^j
\end{equation}
where $\alpha=0.5$ is a trade-off hyperparameter. 
We choose the $m$ features with the highest scores to form the weakly discriminative feature subset $S_e$ of domain $e$.\par

\begin{figure}[!t]
  \centering
   \includegraphics[width=0.8\linewidth]{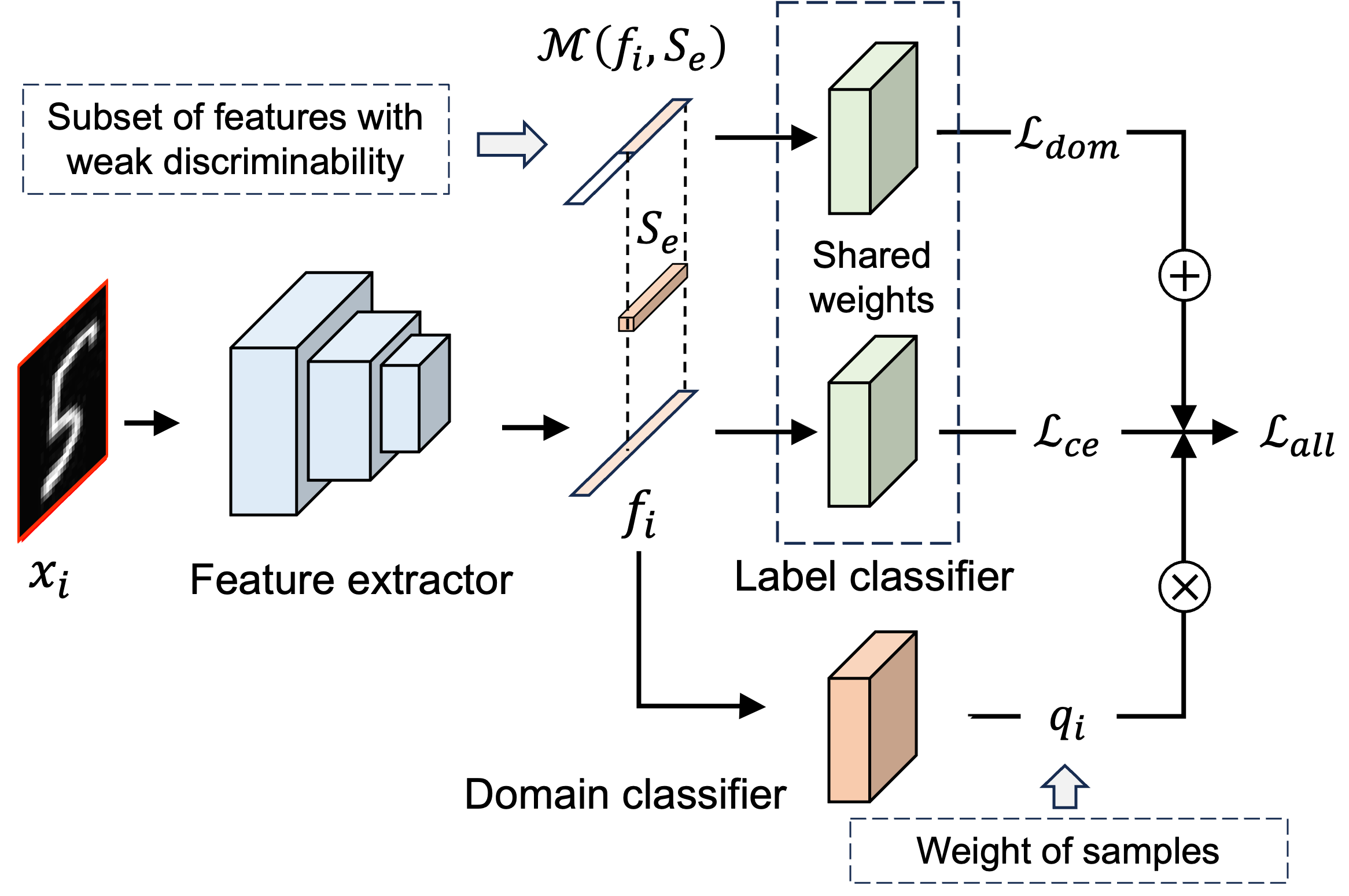}
  \caption{
The calculation of the loss function to optimize weakly discriminative feature subset.
  }
  \label{fig:loss}
\end{figure}

\textbf{Optimizing weakly discriminative features}. 
After identifying the weakly discriminative feature subsets, we adopt a optimization loss during the training phase to increase the inter-class distance.
Fig. \ref{fig:loss} illustrates the process of computing the loss.

We employ the cross entropy function to optimize these weakly discriminative features. 
Given a domain $e$, we utilize the model's classifier to compute the outputs using the features contained in the weak feature subset $S_e$. 

\begin{equation}
\label{eq9}
z_i^\prime=g\left(\mathcal{M}\left(f_i,S_e\right)\right),\ x_i\in X^e
\end{equation}
where $\mathcal{M}\left(\cdot,S_e\right)$ is the mask function that sets the eigenvalues in the non-feature subset to $0$. 
Subsequently, the cross entropy is utilized to compute the classification loss.
To prevent the loss function from only capturing localized features, the SD regular term \cite{DBLP:conf/nips/PezeshkiKBCPL21}  is introduced.
The optimization loss for each sample from domain $e$ is

\begin{equation}
\label{eq10}
\mathcal{L}_{dom}(x_i,y_i)=\mathcal{L}_{ce}\left(z_i^\prime,y_i\right)+\frac{\delta}{2}||z'_i||^2
\end{equation}

Additionally, the premise of this work posits that various samples have distinct effects on the enhancement of generalization ability. 
Consequently, we aim to establish diverse loss weights for each sample. 
We believe that features containing a higher degree of domain-specific information should exert more stringent constraints.
 To quantify the amount of domain information present in each feature, we calculate the posterior probability, denoted as $p\left(e\middle| f_i\right)$.
Upon normalizing the probabilities to have an average value of $1$, they are employed as loss weights, denoted as $q_i$.
Finally, the formula utilized to compute the loss values for each sample is

\begin{equation}
\label{eq11}
\mathcal{L}_{all}(x_i,y_i)=\lambda\mathcal{L}_{ce}\left(x_i,y_i\right)+q_i\mathcal{L}_{dom}(x_i,y_i)
\end{equation}
where $\lambda=\frac{m}{d}$ denotes the decay coefficient employed in the cross-entropy loss across the complete set of features. 
This decay can prevent overfitting resulting from a excessively strong constraint.


\section{Experiments}
In this section, the effectiveness of the DAAL algorithm is validated through experiments conducted on multiple datasets, 
including rotated MNIST \cite{DBLP:conf/iccv/GhifaryKZB15}, Digits \cite{DBLP:conf/cvpr/QiaoZP20}, PACS \cite{DBLP:conf/iccv/LiYSH17}, and VLCS \cite{DBLP:conf/iccv/FangXR13}.
The DAAL algorithm is compared against other domain generalization algorithms as well as classical active learning algorithms. 
Moreover, ablation experiments are conducted to evaluate the individual contributions of the core components.

\subsection{Experimental setting}
As defined in the problem description, the dataset utilized for domain generalization tasks comprises multiple domains, with distinct domains assigned to the training and test sets. 
To accommodate this, we employ the leave-one-out strategy to divide the source and target domains. 
In this strategy, one domain from the dataset is assigned as the test set, while the remaining domains are used to construct the training set.

In this research, we assume that the initial state of the source domain samples is unlabeled. 
Our experiment simulates various levels of labeling capacity to evaluate the model performance across different numbers of labeled samples. 
Specifically, we employ two protocols: 
(1) randomly selecting a specified number of samples for simultaneous labeling, followed by training the model in a supervised manner, 
and (2) employing active learning algorithms to iteratively select samples and train the model.

Next, we provide an introduction to each dataset.
The PACS dataset comprises four domains: art painting (A), cartoon (C),  photo (P), and sketch (S), with a total of 9,991 samples. 
Each domain contains seven categories.
We utilized the training-validation split provided by the dataset \cite{DBLP:conf/iccv/LiYSH17}. 
In the active learning protocol, we labeled 25\% of the training samples in each round, iterating through 4 rounds, including the initialization stage. 
In the random selection protocol, we set the sample sizes to be 50\%, 75\%, and 100\% of the total samples. 
This allows us to evaluate the model's performance under the same number of labels in both protocols.

The VLCS dataset consists of multiple sub-datasets, including PASCAL VOC 2007 (V), LabelMe (L), Caltech (C), and Sun (S), with a total of 10,729 samples. 
Each sub-dataset corresponds to a different domain, and all domains share the same five categories. 
The protocols used for the VLCS dataset were the same as those used for the PACS dataset.

The Digits dataset is a composite digit recognition dataset consisting of MNIST (M), SVHN (S), MNIST-M (MM), SYN, and USPS (U). 
Each sub-dataset has distinct fonts and colors, representing unique domains. 
Due to variations in size among the sub-datasets, we adopted the settings proposed in \cite{DBLP:conf/aaai/ZhouYHX20}, randomly selecting 600 images from each domain for each category. 
In the active learning protocol, we labeled 20\% of the training samples in each of the 5 rounds. 
In the random selection protocol, we employed sample sizes of 40\%, 60\%, 80\%, and 100\%.

The rotated MNIST dataset is a synthetic dataset created by rotating the images from the original MNIST dataset at specific angles, resulting in different domains for each angle. 
We selected specific angles ($0^{\circ}$, $30^{\circ}$, $60^{\circ}$, and $90^{\circ}$), generating four domains.
Each domain contains 50,000 training samples, split into training and validation sets at an 8:2 ratio. 
In the active learning protocol, we labeled 2\% of the training samples in each of the 5 rounds. 
In the random selection protocol,  we used sample sizes of 4\%, 6\%, 8\%, and 10\%.

In the training phase, we employed a pre-trained ResNet18 \cite{DBLP:conf/cvpr/HeZRS16} model on ImageNet as a feature extractor for the PACS and VLCS datasets. 
The PACS dataset was trained with an initial learning rate of 0.004, a batch size of 32, and stochastic gradient descent (SGD) optimizer for 3000 iterations.
For the VLCS dataset, we trained with an initial learning rate of 0.001, a batch size of 64, and a SGD optimizer for 1000 iterations. 
On the other hand, for the Digits and rotated MNIST datasets, we employed a randomly initialized ConvNet \cite{DBLP:conf/iccv/MotiianPAD17} as the feature extractor. 
The training of the Digits dataset employed an initial learning rate of 0.01, a batch size of 64, and an SGD optimizer for 3000 iterations. 
Similarly, the rotated MNIST dataset was trained with an initial learning rate of 0.05, a batch size of 64, and an SGD optimizer for 20000 iterations. 
A cosine scheduler was used for learning rate decay during training.
It is important to note that the number of iterations mentioned pertains to the full training set. 
For the training subset, the training iteration steps are also reduced proportionally.

Regarding the hyperparameter settings, we employed domain adversarial sample selection after using the least confidence method to choose 1.5 times the target number of samples as candidates, denoted as $\rho=1.5$. 
The model training process involved the random forest algorithm with a maximum depth of 5 and 100 trees for calculating feature contributions. 
The size of the weakly discriminative feature subset was set to half of the total number of features, denoted as $\lambda=0.5$ . 
Additionally, we conducted other comparative experiments to maintain consistency with the specific settings of each method. 
To ensure the reliability of the results, we repeated each model's experiment 3 times and reported the average of the results.

\newcolumntype{R}{>{\raggedleft\arraybackslash}p{0.5cm}}

\begin{table}[!t]\footnotesize
\caption{The classification accuracy (\%) of domain generalization methods on the PACS dataset. }
\label{tab:dg_pacs}
\centering
\begin{tabular}{|l|r|r|r|r|r|r|}
\hline
Method                                                        & Scale & \multicolumn{1}{c|}{A} & \multicolumn{1}{c|}{C} & \multicolumn{1}{c|}{P} & \multicolumn{1}{c|}{S} & \multicolumn{1}{c|}{Avg} \\ \hline \hline
ERM \cite{DBLP:journals/tkde/WangLLOQLCZY23}  & 100\%   & 77.00                         & 74.50                        & 95.50                      & 77.80                       & 81.20                    \\ \hline
RSC \cite{DBLP:conf/eccv/HuangWXH20}          & 100\%   & 83.43                         & 80.31                        & 95.99                      & 80.85                       & 85.15                    \\ \hline
SelfReg \cite{DBLP:conf/iccv/KimYPKL21}       & 100\%   & 82.34                         & 78.43                        & 96.22                      & 77.47                       & 83.62                    \\ \hline
CIRL \cite{DBLP:conf/cvpr/LvLLZLWL22}         & 100\%   & \textbf{86.08}                & \textbf{80.59}               & 95.93                      & 82.67                       & \textbf{86.32}           \\ \hline
DFRL \cite{DBLP:journals/isci/WangZZWWL23}    & 100\%   & 85.60                         & 80.10                        & 96.00                      & 79.80                       & 85.38                    \\ \hline
IV-DG \cite{DBLP:journals/tkdd/YuanMXGL0LK23} & 100\%   & 83.36                         & 78.76                        & \textbf{96.87}             & 78.68                       & 84.42                    \\ \hline
NCDG \cite{DBLP:journals/pami/TianLX0023}     & 100\%   & 84.70                         & 77.70                        & 95.40                      & \textbf{82.10}              & 84.98                    \\ \hline
SAGM* \cite{DBLP:conf/cvpr/WangZLZ23}         & 100\%   & 82.28                         & 76.96                        & 95.62                      & 79.46                       & 83.58                    \\ \hline
DAAL ($t$=4)                                              & 100\%   & 80.37                         & 77.60                        & 95.39                      & 78.40                       & 82.94                    \\ \hline \hline
RSC*                                                          & 75\%    & 76.66                         & 73.16                        & 94.95                      & 77.04                       & 80.45                    \\ \hline
CIRL*                                                         & 75\%    & 80.13                         & 76.87                        & 92.39                      & 80.32                       & 82.43                    \\ \hline
SAGM*                                                         & 75\%    & 82.03                         & 74.23                        & 94.07                      & 79.71                       & 82.51                    \\ \hline
DAAL($t$=3)                                                & 75\%    & \textbf{82.71}                & \textbf{78.03}               & \textbf{95.39}             & \textbf{80.50}              & \textbf{84.16}           \\ \hline \hline
RSC*                                                          & 50\%    & 73.73                         & 69.55                        & 94.73                      & 74.29                       & 78.08                    \\ \hline
CIRL*                                                         & 50\%    & 77.30                         & 72.57                        & 91.79                      & 77.24                       & 79.73                    \\ \hline
SAGM*                                                         & 50\%    & \textbf{80.42}                & 71.80                        & 93.89                      & 77.42                       & 80.88                    \\ \hline
DAAL($t$=2)                                              & 50\%    & 80.01                         & \textbf{77.10}               & \textbf{95.20}             & \textbf{80.02}              & \textbf{83.08}           \\ \hline
\end{tabular}
\begin{tablenotes}
        \footnotesize
        \item $\ast$ represents the results of the model that we have reproduced.   
\end{tablenotes}
\end{table}

\newcolumntype{R}{>{\raggedleft\arraybackslash}p{0.5cm}}

\begin{table}[!t]\footnotesize
\caption{The classification accuracy (\%) of domain generalization methods on the VLCS dataset. }
\label{tab:dg_vlcs}
\centering
\begin{tabular}{|l|r|r|r|r|r|r|}
\hline
Method     & Scale & \multicolumn{1}{c|}{V} & \multicolumn{1}{c|}{L} & \multicolumn{1}{c|}{C} & \multicolumn{1}{c|}{S} & \multicolumn{1}{c|}{Avg} \\ \hline\hline
ERM*       & 100\%                      & 70.71                                         & 60.73                  & 97.48                  & 67.17                                         & 74.02                    \\ \hline
RSC \cite{DBLP:conf/eccv/HuangWXH20}         & 100\%                      & 73.30                                         & 61.86                  & \textbf{97.61}         & 68.32                                         & 75.27                    \\ \hline
MMLD \cite{DBLP:conf/aaai/MatsuuraH20}       & 100\%                      & 71.96                                         & 58.77                  & 96.66                  & 68.13                                         & 73.88                    \\ \hline
MFE \cite{DBLP:journals/pr/ZhangZWL23a}       & 100\%                      & 71.00                                         & 62.70                  & 97.60                  & 69.60                                         & 75.23                    \\ \hline
NCDG \cite{DBLP:journals/pami/TianLX0023}      & 100\%                      & 70.70                                         & \textbf{67.60}         & 97.20                  & 68.70                                         & 76.05                    \\ \hline
SAGM*\cite{DBLP:conf/cvpr/WangZLZ23}     & 100\%                      & 73.37                                         & 62.45                  & 96.57                  & 71.65                                         & 76.01                    \\ \hline
DAAL($t$=4) & 100\%                      & \textbf{74.82}                                & 60.98                  & 97.17                  & \textbf{71.83}                                & \textbf{76.20}           \\ \hline\hline
ERM*       & 75\%                       & 70.55                                         & 59.60                  & \textbf{97.58}         & 68.74                                         & 74.12                    \\ \hline
RSC*       & 75\%                       & 72.54                                         & 60.41                  & 97.47                  & 70.10                                         & 75.13                    \\ \hline
SAGM*      & 75\%                       & 74.37                                         & 59.60                  & 97.17                  & 71.70                                         & 75.71                    \\ \hline
DAAL($t$=3) & 75\%                       & \textbf{75.71}                                & \textbf{62.23}         & 97.17                  & \textbf{71.90}                                & \textbf{76.75}           \\ \hline\hline
ERM*       & 50\%                       & 70.94                                         & 58.68                  & 96.06                  & 66.82                                         & 73.125                   \\ \hline
RSC*       & 50\%                       & 70.89                                         & 60.19                  & \textbf{97.57}         & 66.74                                         & 73.85                    \\ \hline
SAGM*      & 50\%                       & 73.30                                         & 57.40                  & 96.57                  & \textbf{70.35}                                & 74.41                    \\ \hline
DAAL ($t$=2)  & 50\%                       & \textbf{75.21}                                & \textbf{61.36}         & 97.41                  & 68.92                                & \textbf{75.73}           \\ \hline
\end{tabular}
\end{table}

\subsection{Comparison with domain generalization}
We compared our DAAL algorithm with other mainstream domain generalization algorithms on the PACS, VLCS, Digits, and rotated MNIST datasets. 
To present the performance of models under different sample scales, we reproduced some of the comparison algorithms.
The results of the DAAL algorithm after each round is compared with the results of the comparison models based on the same sample size following the random selection protocol.

\textbf{Evaluation on the PACS dataset.}
We compared the DAAL algorithm with other algorithms such as Empirical Risk Minimization (ERM) \cite{DBLP:journals/tkde/WangLLOQLCZY23} , RSC \cite{DBLP:conf/eccv/HuangWXH20}, SelfReg \cite{DBLP:conf/iccv/KimYPKL21}, CIRL \cite{DBLP:conf/cvpr/LvLLZLWL22}, DFRL \cite{DBLP:journals/isci/WangZZWWL23}, IV-DG \cite{DBLP:journals/tkdd/YuanMXGL0LK23}, NCDG \cite{DBLP:journals/pami/TianLX0023}, and SAGM \cite{DBLP:conf/cvpr/WangZLZ23} . 
To assess the performance at different sample scales (50\% and 75\%), we reproduced the RSC, CIRL, and SAGM algorithms using randomly sampled data. 
Tab. \ref{tab:dg_pacs} shows the comparison experiment results on the PACS dataset. 
When considering a 100\% training sample scale, the DAAL algorithm did not exhibit a significant advantage. 
However, with a reduced training sample scale of 75\%, the DAAL algorithm achieved an average accuracy of 84.15\% across all four domains. 
Notably, when focusing on the cartoon and sketch domains, our model's performance even approached that of some state-of-the-art models using a full training sample size. 
On the other hand, the performance of the RSC, CIRL, and SAGM models declined significantly due to the reduction in training samples.
At a sample scale of 50\%, the impact on the comparison models was more pronounced, while the DAAL algorithm maintained an average accuracy of 83.08\%, demonstrating a clear advantage. 
These results underscore the substantial influence of the number of training samples on the model's performance in the PACS dataset. 
Our DAAL algorithm showcases the ability to improve model generalization under labeling-capability-limited conditions.
Regarding the issue of decreased model performance of the DAAL algorithm with a 100\% training sample scale, we attribute it to distribution shift. 
The high performance observed on the source domain validation sets may indicate a decline in performance on the target domain test set.


\begin{table}[!t]\footnotesize
\caption{The classification accuracy (\%) of domain generalization methods on the Digits dataset. }
\label{tab:dg_digit}
\centering
\begin{tabular}{|l|r|r|r|r|r|r|r|}
\hline
Method & Scale & \multicolumn{1}{c|}{M} & \multicolumn{1}{c|}{MM} & \multicolumn{1}{c|}{S} & \multicolumn{1}{c|}{SYN} & \multicolumn{1}{c|}{U} & \multicolumn{1}{c|}{Avg} \\ \hline\hline
ERM    & 100\% & 97.57                      & 68.33                        & 73.61                     & 79.12                    & 93.76                     & 84.36                    \\ \hline
IRM\cite{DBLP:journals/corr/abs-1907-02893}    & 100\% & 97.66                      & 68.92                        & 74.31                     & 79.15                    & 93.72                     & 84.58                    \\ \hline
RSC    & 100\% & 97.65                      & 69.06                        & 75.47                     & 79.29                    & 93.66                     & 84.80                    \\ \hline
SAGM   & 100\% & 97.63                      & \textbf{69.58}               & 74.41                     & 79.29                    & 93.60                     & 84.69                    \\ \hline
DAAL($t$=5)   & 100\% & \textbf{97.82}             & 68.69                        & \textbf{75.50}            & \textbf{79.97}           & \textbf{95.05}            & \textbf{85.35}           \\ \hline\hline
ERM    & 80\%  & 97.50                      & 68.33                        & 72.39                     & 78.24                    & 93.06                     & 83.76                    \\ \hline
IRM    & 80\%  & 97.52                      & 68.76                        & 73.01                     & 78.55                    & 93.12                     & 84.01                    \\ \hline
RSC    & 80\%  & 97.42                      & \textbf{69.44}               & 74.00                     & 78.83                    & 93.11                     & 84.32                    \\ \hline
SAGM   & 80\%  & 97.32                      & 68.77                        & 72.98                     & 78.50                    & 93.17                     & 83.99                    \\ \hline
DAAL($t$=4)   & 80\%  & \textbf{97.80}             & 68.81                        & \textbf{74.82}            & \textbf{79.73}           & \textbf{95.75}            & \textbf{85.44}           \\ \hline\hline
ERM    & 60\%  & 96.95                      & 67.61                        & 70.31                     & 76.96                    & 93.65                     & 83.19                    \\ \hline
IRM    & 60\%  & 96.57                      & 67.99                        & 70.95                     & 77.55                    & 94.02                     & 83.52                    \\ \hline
RSC    & 60\%  & 96.56                      & 68.77                        & 71.49                     & 77.58                    & \textbf{94.10}            & 83.77                    \\ \hline
SAGM   & 60\%  & 97.10                      & 68.75                        & 71.03                     & 77.63                    & 93.92                     & 83.73                    \\ \hline
DAAL($t$=3)   & 60\%  & \textbf{97.47}             & \textbf{68.93}               & \textbf{75.10}            & \textbf{79.86}           & 93.98                     & \textbf{84.89}           \\ \hline\hline
ERM    & 40\%  & 96.09                      & \textbf{68.40}               & 69.12                     & 75.09                    & 93.35                     & 82.57                    \\ \hline
IRM    & 40\%  & 96.08                      & 67.66                        & 69.03                     & 75.20                    & 93.75                     & 82.58                    \\ \hline
RSC    & 40\%  & 95.97                      & 67.52                        & 69.05                     & 73.71                    & 93.66                     & 82.26                    \\ \hline
SAGM   & 40\%  & 96.22                      & 67.46                        & 69.68                     & 74.49                    & 93.60                     & 82.51                    \\ \hline
DAAL($t$=2)   & 40\%  & \textbf{96.40}             & 67.76                        & \textbf{72.28}            & \textbf{78.40}           & \textbf{93.90}            & \textbf{83.77}           \\ \hline
\end{tabular}
\end{table}

\textbf{Evaluation on the VLCS dataset.}
we adopt ERM, RSC, MMLD \cite{DBLP:conf/aaai/MatsuuraH20}, MFE \cite{DBLP:journals/pr/ZhangZWL23a}, NCDG, SAGM as comparison algorithm. 
The experimental results are shown in Tab. \ref{tab:dg_vlcs} . 
With a 100\% training sample scale, the DAAL algorithm demonstrated comparable performance to other algorithms, achieving an average accuracy of 76.2\%. 
When the training sample scale was reduced to 75\% and 50\%, the other algorithms exhibited a noticeable decrease in accuracy. 
In contrast, the DAAL algorithm achieved higher accuracy with the same quantity of samples, indicating its advantage over other algorithms.
These findings highlight the effectiveness of the sampling strategy employed by active learning algorithms to enhance model training performance in datasets with imbalanced sample distributions.


\begin{table}[!t]\footnotesize
\caption{The classification accuracy (\%) of domain generalization methods on the rotated MNIST dataset. }
\label{tab:dg_rmnist}
\centering
\begin{tabular}{|l|r|r|r|r|r|r|}
\hline
Method & Scale & \multicolumn{1}{c|}{$0^{\circ}$} & \multicolumn{1}{c|}{$30^{\circ}$} & \multicolumn{1}{c|}{$60^{\circ}$} & \multicolumn{1}{c|}{$90^{\circ}$} & \multicolumn{1}{c|}{Avg} \\ \hline\hline
ERM    & 10\%  & 75.06                  & 94.32                   & 94.34                   & 77.01                   & 85.18                    \\ \hline
IRM    & 10\%  & 75.50                  & 94.80                   & 95.30                   & 76.99                   & 85.65                    \\ \hline
RSC    & 10\%  & 75.30                  & 93.87                   & 94.96                   & 76.87                   & 85.25                    \\ \hline
SAGM   & 10\%  & 75.55                  & 93.50                   & 94.21                   & 76.51                   & 84.94                    \\ \hline
DAAL($t$=5)   & 10\%  & \textbf{80.83}         & \textbf{96.85}          & \textbf{97.12}          & \textbf{89.38}          & \textbf{91.05}           \\ \hline\hline
ERM    & 8\%   & 72.94                  & 93.89                   & 94.01                   & 75.19                   & 84.01                    \\ \hline
IRM    & 8\%   & 74.80                  & 94.25                   & 94.56                   & 75.20                   & 84.70                    \\ \hline
RSC    & 8\%   & 74.60                  & 94.08                   & 94.23                   & 75.63                   & 84.64                    \\ \hline
SAGM   & 8\%   & 72.64                  & 94.48                   & 94.23                   & 75.01                   & 84.09                    \\ \hline
DAAL($t$=4)   & 8\%   & \textbf{79.78}         & \textbf{96.43}          & \textbf{96.68}          & \textbf{87.97}          & \textbf{90.22}           \\ \hline\hline
ERM    & 6\%   & 71.79                  & 93.01                   & 93.52                   & 74.16                   & 83.12                    \\ \hline
IRM    & 6\%   & 72.05                  & 93.56                   & 94.11                   & 74.40                   & 83.53                    \\ \hline
RCS    & 6\%   & 72.72                  & 93.38                   & 94.05                   & 72.62                   & 83.19                    \\ \hline
SAGM   & 6\%   & 70.47                  & 93.90                   & 93.46                   & 74.47                   & 83.08                    \\ \hline
DAAL($t$=3)   & 6\%   & \textbf{78.35}         & \textbf{96.10}          & \textbf{96.61}          & \textbf{87.37}          & \textbf{89.61}           \\ \hline\hline
ERM    & 4\%   & 69.68                  & 92.86                   & 92.54                   & 71.90                   & 81.75                    \\ \hline
IRM    & 4\%   & 70.11                  & 92.45                   & 92.98                   & 71.77                   & 81.83                    \\ \hline
RSC    & 4\%   & 71.41                  & 93.29                   & 93.31                   & 71.79                   & 82.45                    \\ \hline
SAGM   & 4\%   & 68.76                  & 92.98                   & 92.04                   & 70.33                   & 81.03                    \\ \hline
DAAL($t$=2)   & 4\%   & \textbf{79.90}         & \textbf{94.81}          & \textbf{95.71}          & \textbf{84.11}          & \textbf{88.63}           \\ \hline
\end{tabular}
\end{table}

\textbf{Evaluation on the Digits dataset.}
We reproduced ERM, IRM \cite{DBLP:journals/corr/abs-1907-02893}, RSC, and SAGM algorithms to evaluate their performances at varing sample scales. 
The experimental results can be found in Tab. \ref{tab:dg_digit}. 
When employing MNIST, MNIST-M, and USPS as target domains, the models demonstrated relatively low sensitivity to changes in the training sample scale. 
For instance, the ERM model exhibited only a slight 1\% difference in results between using 40\% and 100\% training samples. 
Despite this, our DAAL algorithm still maintained a slight advantage in these three domains.
On the other hand, when considering SVHN and SYN as target domains, the accuracy of the ERM model declined by approximately 4\% due to the reduction in the sample quantity. 
The DAAL algorithm achieved comparable performance even when trained on only 60\% of the samples, matching the performance of the fully sampled model.
These findings highlight the capability of the DAAL algorithm to enhance domain generalization performance, particularly in scenarios with limited sample sizes.

\textbf{Evaluation on the rotated MNIST dataset.}  
We reproduced ERM, IRM, RSC, and SAGM algorithms on the rotated MNIST dataset. 
Given the ample training data available for MNIST and its relatively low recognition difficulty, we limited the sample usage to a maximum of 10\%.
In the experimental results shown in Tab. \ref{tab:dg_rmnist}, when the target domains were set with angles of $30^{\circ}$ and $60^{\circ}$, the models achieved test set accuracies exceeding 90\%. 
The DAAL algorithm consistently performed well across different training sample scales, exhibiting an accuracy improvement of 2\%-3\%. 
However, when the target domains were set at $0^{\circ}$ and $90^{\circ}$, representing the extreme angles, the models encountered greater learning challenges. 
In these tasks, our DAAL algorithm showcased superior performance, achieving an accuracy improvement exceeding 10\% compared to the control group with just 4\% of the training samples utilized. 
These findings indicate that the samples chosen by our DAAL algorithm possess higher learning value than the randomly selected samples used by other algorithms. 
This emphasizes the effectiveness of our DAAL algorithm, especially in challenging domains where there is limited labeling capacity.

\subsection{Comparison with active learning}
To further validate the contribution of our approach to active learning, we compared it with classical active learning algorithms such as the least confidence (LeastConf) \cite{DBLP:conf/ijcnn/WangS14}, the maximum entropy (Entropy) \cite{2010Active}, 
BALD \cite{DBLP:conf/icml/GalIG17}, and BADGE \cite{DBLP:conf/iclr/AshZK0A20} methods. 
In the comparative experiments, we employed the RSC model as the training model and evaluated its performance on target domains that are sensitive to the scale of training samples.

\begin{figure}[!t]
  \centering
  \subfigure[Artpainting]{
   \includegraphics[width=0.45\linewidth]{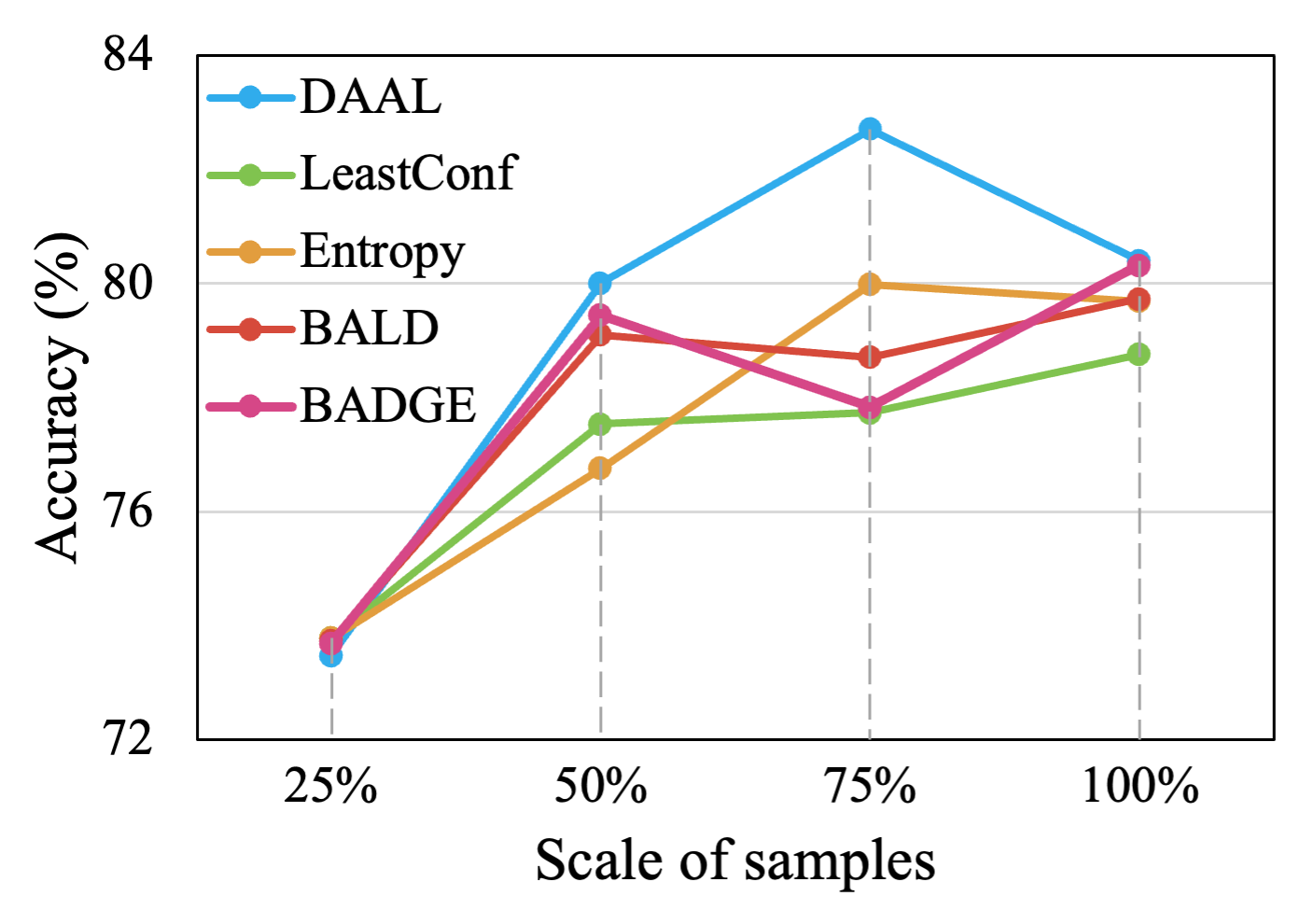}
  }
 \subfigure[Cartoon]{
   \includegraphics[width=0.45\linewidth]{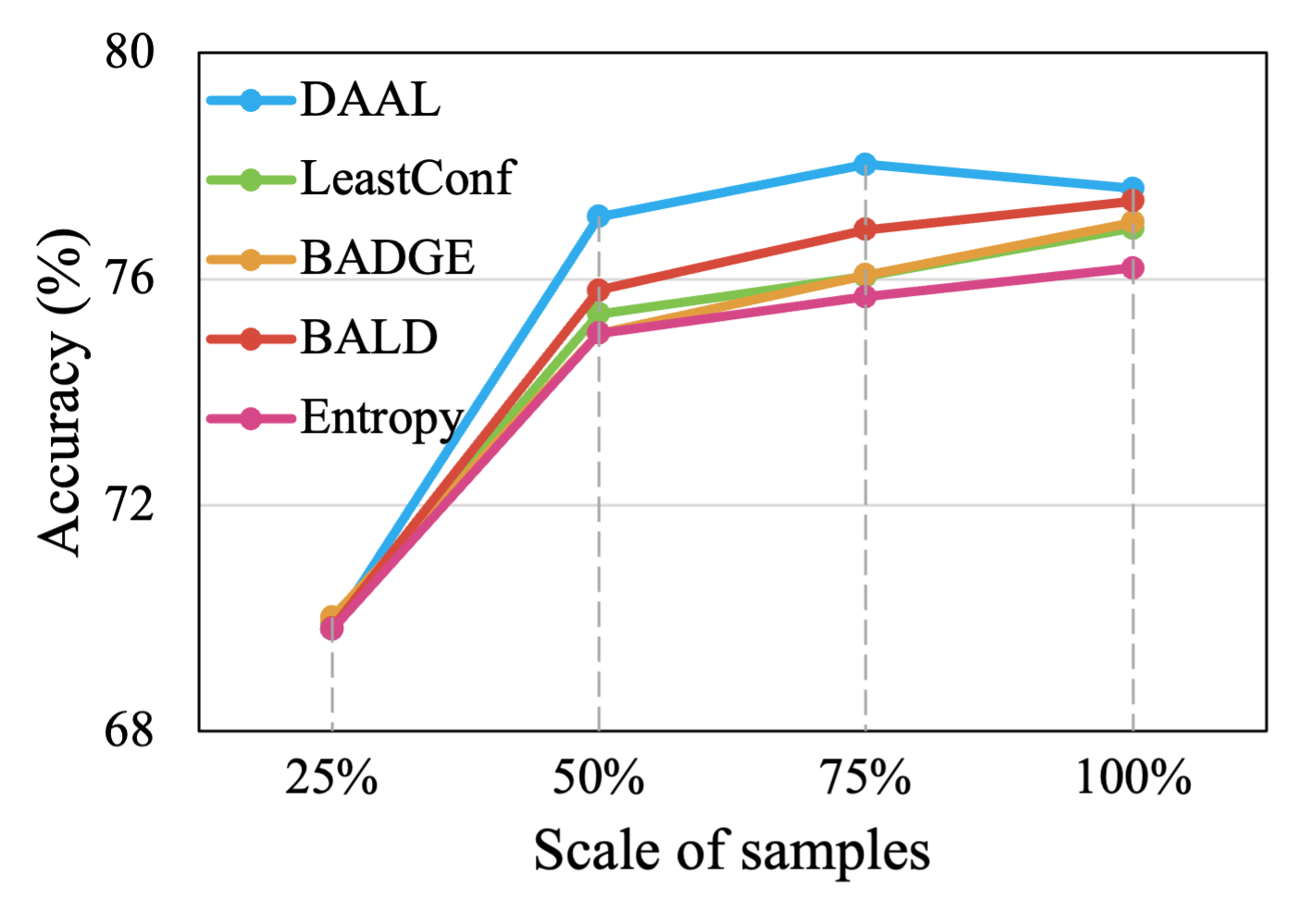}
  }
  \caption{
The results of active learning in the PACS dataset. 
  }
  \label{fig:al_pacs}
\end{figure}

\textbf{Evaluation on the PACS dataset.}
The generalization capability is closely related to the scale of training samples, particularly when dealing with art painting and cartoon domains as target domains.
Therefore, we conducted active learning comparative experiments in these two domains. 
The results, as shown in the Fig. \ref{fig:al_pacs}, reveal that the DAAL algorithm consistently achieves significantly higher test accuracies after the 2nd and 3rd iterations, as compared to other active learning algorithms. 
This outcome underscores the ability of the DAAL algorithm to maximize model generalization by effectively selecting appropriate samples, and integrating specific optimization techniques. 
Furthermore, as we moved to the 4th iteration where the full dataset was utilized, the performance differences between the algorithms diminished. 
This reduction can be attributed to the absence of difference on the training samples.

\begin{figure}[!t]
  \centering
  \subfigure[SVHN]{
   \includegraphics[width=0.45\linewidth]{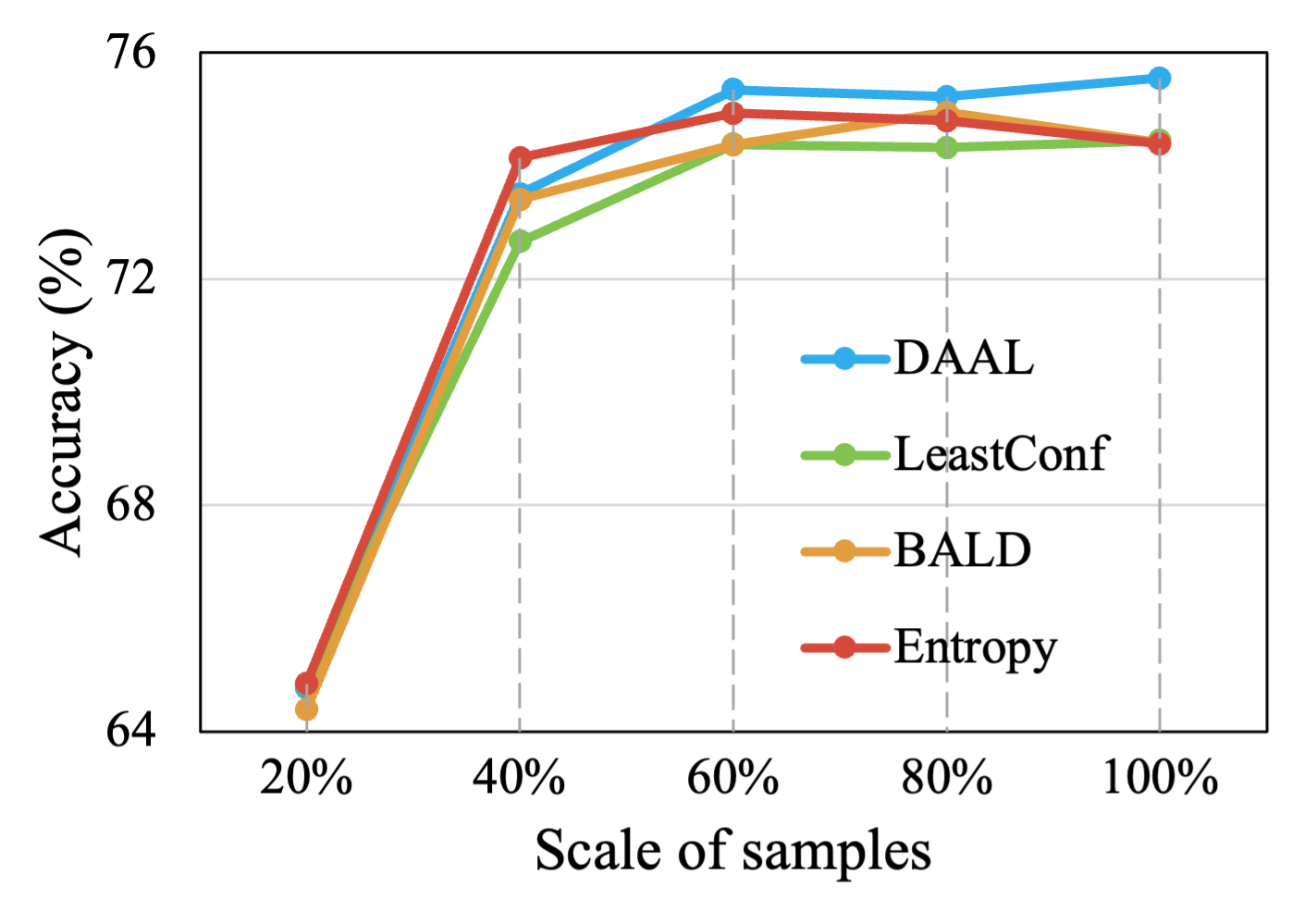}
  }
 \subfigure[SYN]{
   \includegraphics[width=0.45\linewidth]{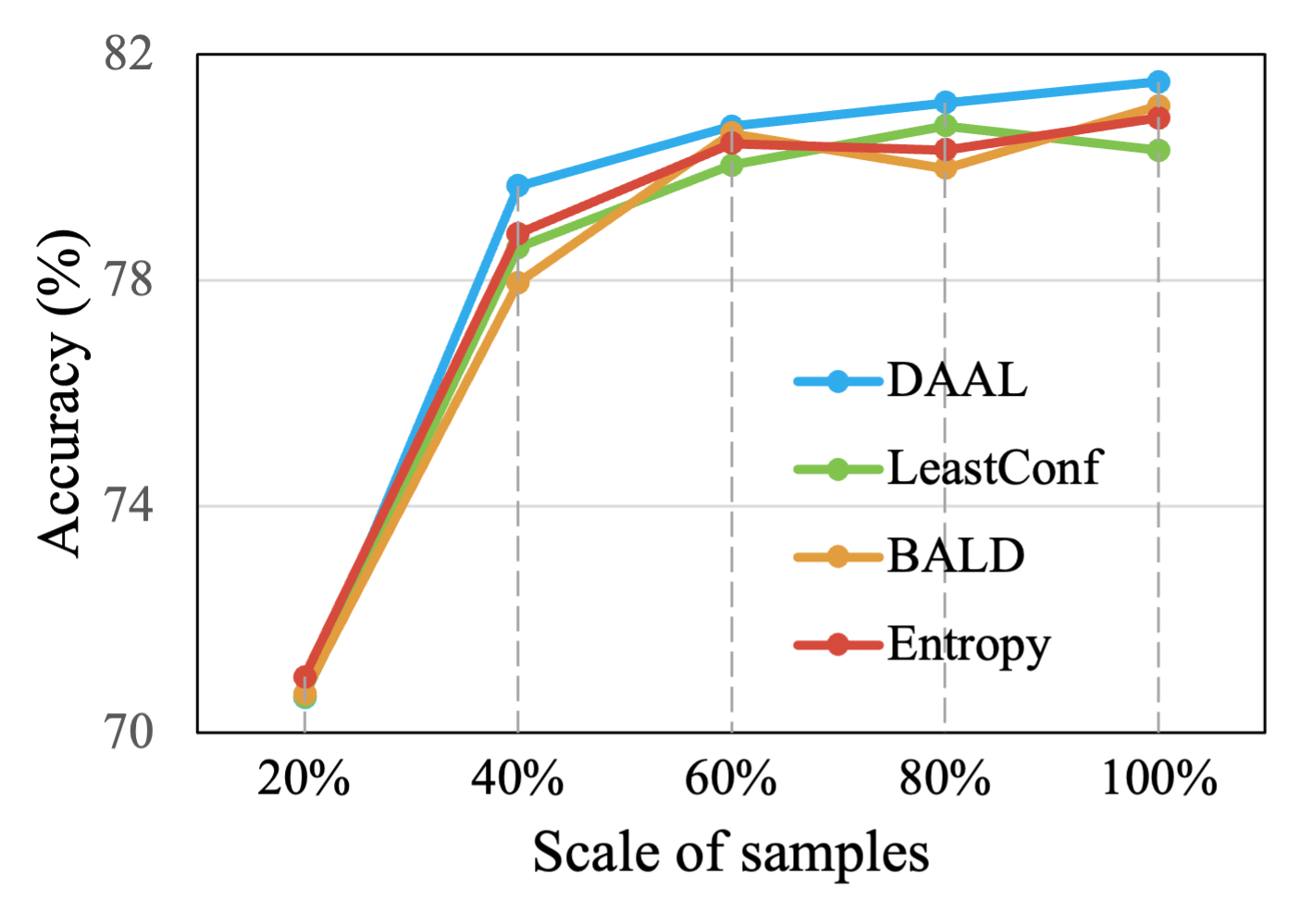}
  }
  \caption{
The results of active learning in the Digits dataset. 
  }
  \label{fig:al_digit}
\end{figure}

\textbf{Evaluation on the Digits dataset.}
The SVHN and SYN domains within the Digits dataset demonstrate greater sensitivity to the scale of training samples.
We conducted experiments on these two domains as target domains. 
The results are displayed in the Fig. \ref{fig:al_digit}.
In the validation of the SVHN domain, the DAAL algorithm consistently achieved higher test results after the 3rd iteration and maintained its advantage throughout subsequent iterations. 
Similar trends can also be observed in the SYN domain.
Conversely, we noticed that some comparative algorithms exhibited a decrease in model performance as the number of samples increased. 
We attribute this phenomenon to distribution shift in the test set.
However, the DAAL algorithm did not encounter such issues in these two domains, indicating its capability to mitigate distribution shift problems to some extent.

\begin{figure}[!t]
  \centering
  \subfigure[$0^{\circ}$ rotation]{
   \includegraphics[width=0.45\linewidth]{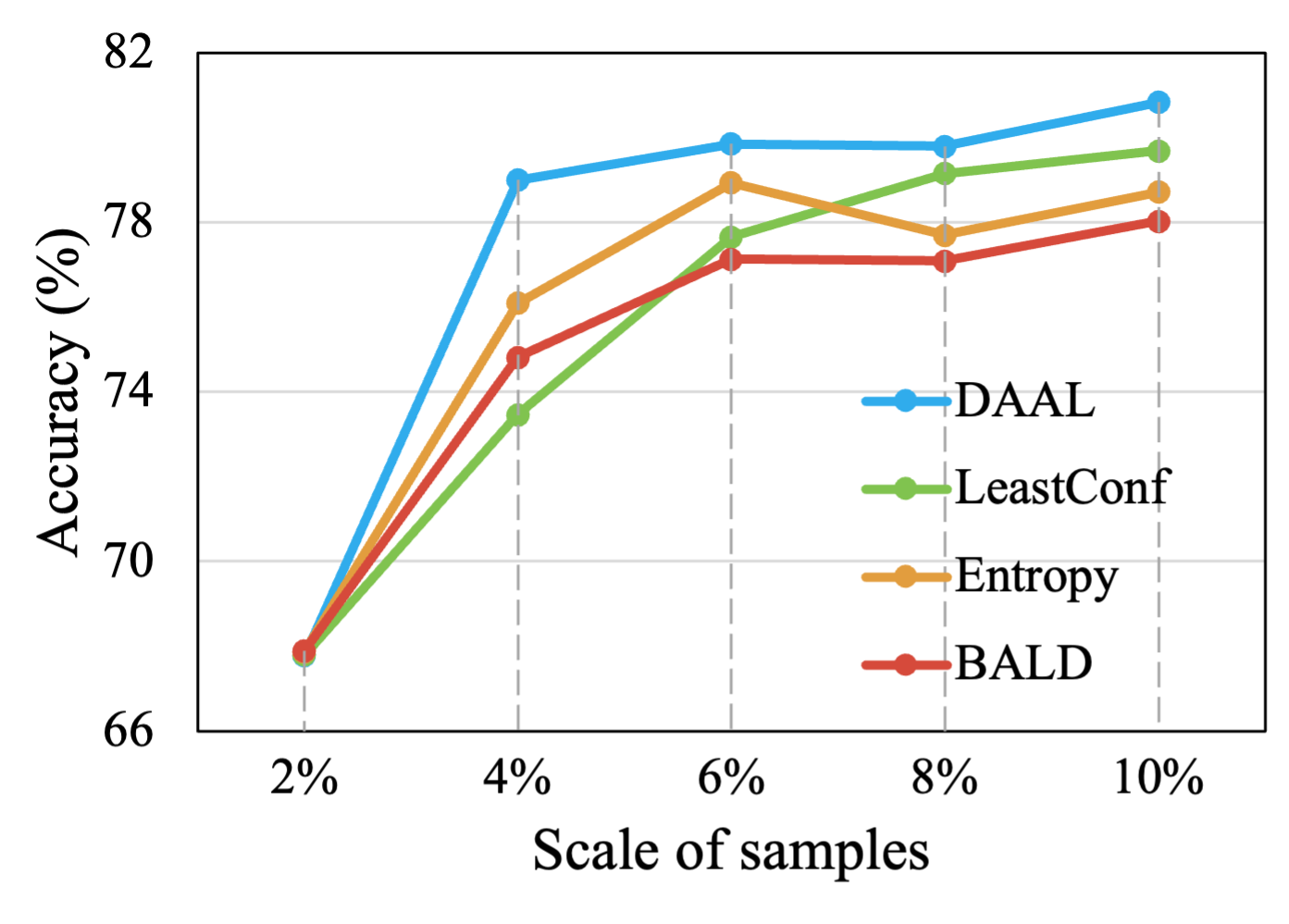}
  }
 \subfigure[$90^{\circ}$ rotation]{
   \includegraphics[width=0.45\linewidth]{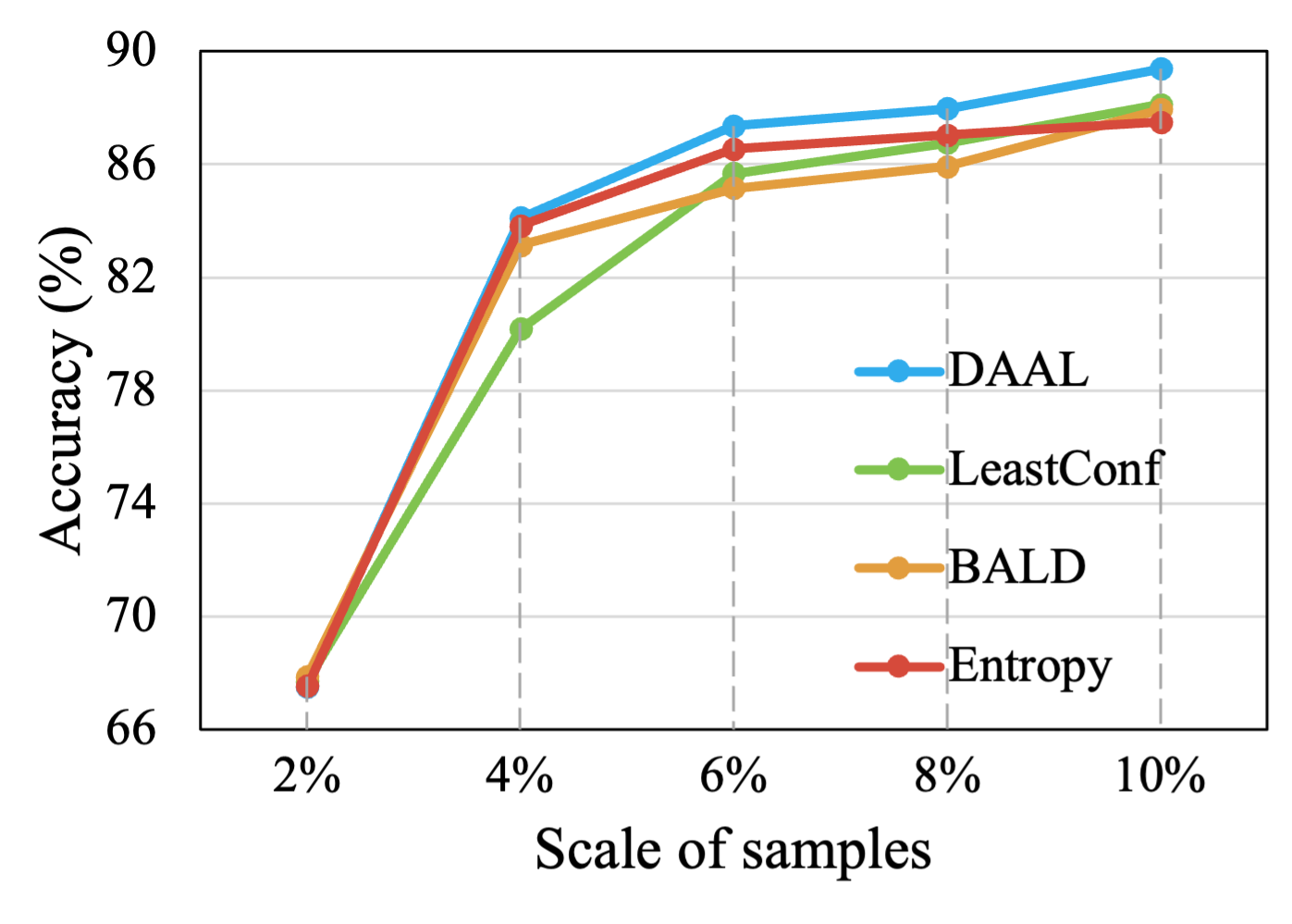}
  }
  \caption{
The results of active learning in the rotated MNIST dataset. 
  }
  \label{fig:al_rmnist}
\end{figure}

\textbf{Evaluation on the rotated MNIST  dataset.}
We selected the $0^{\circ}$ and $90^{\circ}$ as target domains to conduct experiments. 
The results are depicted in the Fig. \ref{fig:al_rmnist}. 
Given the large pool of candidate samples and the relatively low degree of task difficulty in the Rotated MNIST dataset, 
The DAAL algorithm consistently demonstrated advantages over other comparative algorithms during the active learning iteration process. 
Under the same labeling conditions, the DAAL algorithm maintained higher accuracy. 
These findings suggest that when an large number of candidate samples is available, the utilization of the DAAL algorithm for selecting domain adversarial samples proves more advantageous in enhancing the domain generalization ability.

\subsection{Ablation learning}
We conducted ablation experiments on the PACS dataset, as the benefits of the DAAL algorithm are more evident in the datasets.
Our  DAAL algorithm involve selecting domain adversarial samples and optimizing weakly discriminative feature. 
We will conduct separate analyses on both aspects.

\textbf{Selecting domain adversarial samples.}
The primary objective is to select challenging samples that maximize the intra-class distance across different domains while minimizing the inter-class distance within the same domains. 
Correspondingly, we have designed two metric functions (See Eq. \ref{eq5} and \ref{eq6}).
We prepared four groups of experiments.
The first and fundamental group does not utilize any of the metrics but instead directly applies the LeastConf method.
The other two groups utilize one of the metrics individually to select challenging samples. 
The final group combines the results of both metrics, as introduced in our method.

\begin{table}[!t]\footnotesize
\caption{The results of the ablation experiments for selecting domain adversarial samples on the PACS dataset.}
\label{tab:ab_pacs_sample}
\centering
\begin{tabular}{|l|c|c|r|r|r|r|r|}
\hline
Scale                                            & $\varphi_{inter}^{same}$  & $\varphi_{intra}^{cross}$ & \multicolumn{1}{c|}{A} & \multicolumn{1}{c|}{C} & \multicolumn{1}{c|}{P} & \multicolumn{1}{c|}{S} & \multicolumn{1}{c|}{Avg} \\ \hline\hline
\multirow{4}{*}{\begin{tabular}[c]{@{}l@{}}100\%\\ ($t$=4)\end{tabular}} & $\times$ & $\times$ & 80.22                  & 76.62                  & 95.69                  & 79.91                  & 83.11                    \\ \cline{2-8} 
                                                                       & $\surd$ & $\times$ & 80.46                  & 77.34                  & 95.03                  & \textbf{80.38}                  & \textbf{83.30}                    \\ \cline{2-8} 
                                                                       & $\times$ & $\surd$ & \textbf{80.71}                  & 76.45                  & \textbf{95.99}                  & 78.72                  & 82.97                    \\ \cline{2-8} 
                                                                       & $\surd$ & $\surd$ & 80.37                  & \textbf{77.60}                 & 95.39                  & 78.40                  & 82.94                    \\ \hline\hline
\multirow{4}{*}{\begin{tabular}[c]{@{}l@{}}75\%\\ ($t$=3)\end{tabular}}  & $\times$ & $\times$ & 81.74                  & 76.83                  & \textbf{95.87}                  & 78.82                  & 83.32                    \\ \cline{2-8} 
                                                                       & $\surd$ & $\times$ & 82.17                  & 76.41                  & 95.39                  & 77.96                  & 82.98                    \\ \cline{2-8} 
                                                                       & $\times$ & $\surd$ & 79.05                  & \textbf{78.15}                  & 95.39                  & 77.99                  & 82.64                    \\ \cline{2-8} 
                                                                       & $\surd$ & $\surd$ & \textbf{82.71}                  & 78.03                  & 95.39                  & \textbf{80.50}                  & \textbf{84.16}                    \\ \hline\hline
\multirow{4}{*}{\begin{tabular}[c]{@{}l@{}}50\%\\ ($t$=2)\end{tabular}}  & $\times$ & $\times$ & 80.22                  & 74.32                  & \textbf{96.11}                  & 79.00                  & 82.41                    \\ \cline{2-8} 
                                                                       & $\surd$ & $\times$ & \textbf{81.49}                  & 74.74                  & 95.69                  & 76.30                  & 82.06                    \\ \cline{2-8} 
                                                                       & $\times$ & $\surd$ & 77.98                  & 76.32                  & 95.63                  & 79.56                  & 82.37                    \\ \cline{2-8} 
                                                                       & $\surd$ & $\surd$ & 80.01                  & \textbf{77.10}                  & 95.21                  & \textbf{80.02}                  & \textbf{83.08}                    \\ \hline
\end{tabular}
\end{table}

The results are shown in Tab. \ref{tab:ab_pacs_sample}.
After the 2nd and 3rd iterations, the final group that combine the two metrics demonstrates improved generalization ability.
The average accuracy across the four domains increased by nearly 1\%.
With the 4th iteration, when the full dataset is utilized and no differences exist in the training data, the performance of each group of models becomes similar.


\textbf{Optimizing weakly discriminative features.}
In the model training process, our DAAL algorithm incorporates an optimized loss function for weakly discriminative feature subsets in each domain, and assigns varying weights to the optimization loss for individual samples.
Therefore, 3 groups of experiments were designed to investigate the aforementioned aspects.
The first group only uses the cross-entropy function for training. 
The second group adds extra optimized loss based on feature subsets, but assigning a uniform loss weight of 1 to each sample.
The final group builds upon the second group and assigns weights to the loss.

\begin{table}[!t]\footnotesize
\caption{The results of the ablation experiments for optimizing weakly discriminative features on the PACS dataset.}
\label{tab:ab_pacs_feature}
\centering
\begin{tabular}{|l|l|c|r|r|r|r|r|}
\hline
Scale                                                                  & \multicolumn{1}{c|}{Loss} & Weight & \multicolumn{1}{c|}{A} & \multicolumn{1}{c|}{C} & \multicolumn{1}{c|}{P} & \multicolumn{1}{c|}{S} & \multicolumn{1}{c|}{Avg} \\ \hline\hline
\multirow{3}{*}{\begin{tabular}[c]{@{}l@{}}100\%\\ ($t$=4)\end{tabular}} & $\times$                         & $\times$      & 78.51                  & 76.39                  & 95.15                  & 79.69                  & 82.43                    \\ \cline{2-8} 
                                                                       & $\surd$                         & $\times$      & \textbf{81.78}         & 76.66                  & \textbf{95.69}         & \textbf{79.69}         & \textbf{83.46}           \\ \cline{2-8} 
                                                                       & $\surd$                         & $\surd$      & 80.37                  & \textbf{77.60}         & 95.39                  & 78.40                  & 82.94                    \\ \hline\hline
\multirow{3}{*}{\begin{tabular}[c]{@{}l@{}}75\%\\ ($t$=3)\end{tabular}}  & $\times$                         & $\times$      & 79.98                  & 77.73                  & 95.09                  & 79.68                  & 83.12                    \\ \cline{2-8} 
                                                                       & $\surd$                         & $\times$      & 82.13                  & \textbf{78.63}         & \textbf{95.45}                  & 79.94                  & 84.04                    \\ \cline{2-8} 
                                                                       & $\surd$                         & $\surd$      & \textbf{82.71}         & 78.03                  & 95.39                  & \textbf{80.50}         & \textbf{84.16}           \\ \hline\hline
\multirow{3}{*}{\begin{tabular}[c]{@{}l@{}}50\%\\ ($t$=2)\end{tabular}}  & $\times$                         & $\times$      & 79.25                  & 75.94                  & 95.21                  & 74.45                  & 81.21                    \\ \cline{2-8} 
                                                                       & $\surd$                         & $\times$      & 77.59                  & 76.62                  & \textbf{95.27}         & 74.96                  & 81.11                    \\ \cline{2-8} 
                                                                       & $\surd$                         & $\surd$      & \textbf{80.01}         & \textbf{77.10}         & 95.20                  & \textbf{80.02}         & \textbf{83.08}           \\ \hline
\end{tabular}
\end{table}

The experimental results are shown in Tab. \ref{tab:ab_pacs_feature}. 
Following the 2nd iteration, the final group exhibits noticeable performance advantages in terms of accuracy compared to the other two groups. 
Upon reaching the 3rd iteration and being influenced by a larger sample size, all groups of experiments demonstrate improved model performance. 
While the model results of the final group are comparable to the second group without added weights, they still outperform the first group that lacks the inclusion of loss. 
However, after the 4th iteration, there is a certain decline in performance across all groups of models, which can be attributed to distribution bias as previously discussed.


\begin{figure}[!t]
  \centering
  \subfigure[Baseline]{
   \includegraphics[height=1.2in]{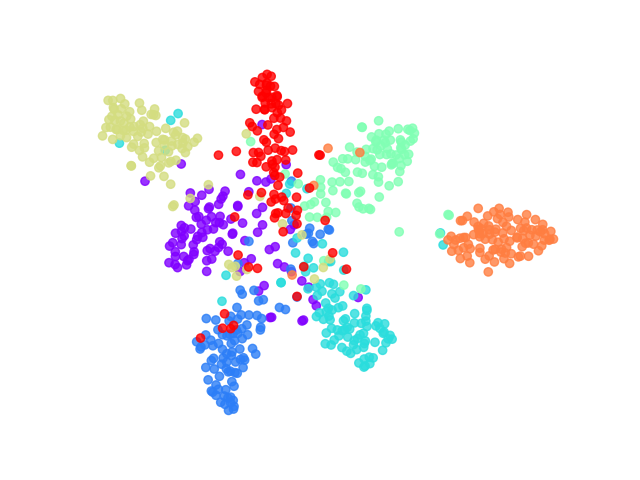}
  }
 \subfigure[Sample selection]{
   \includegraphics[height=1.2in]{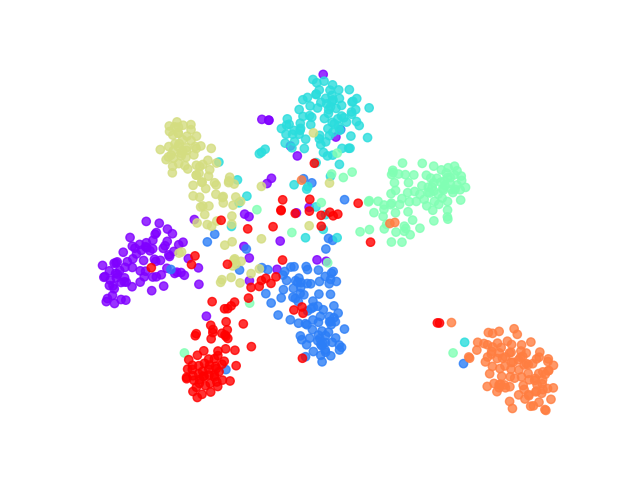}
  }
   \subfigure[Feature optimization]{
   \includegraphics[height=1.2in]{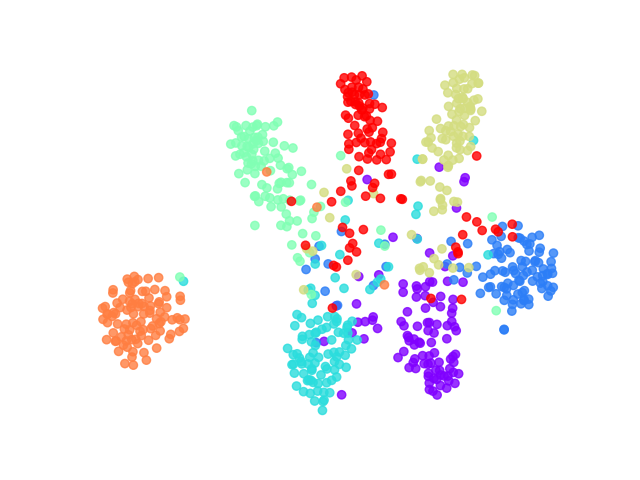}
  }
   \subfigure[Sample selection + Feature optimization]{
   \includegraphics[height=1.2in]{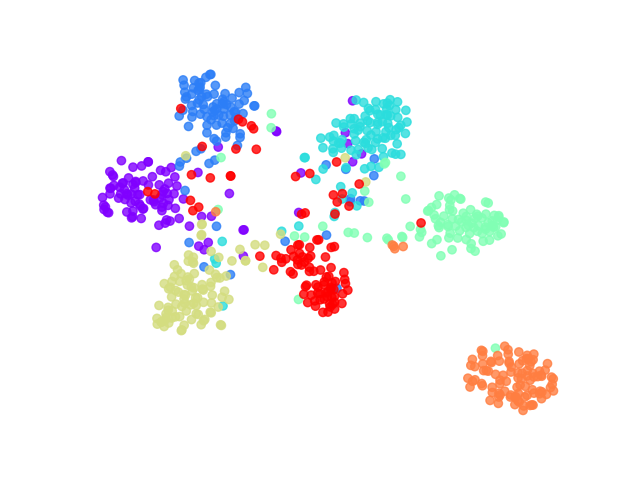}
  }
  \caption{
T-SNE Visualization results of ablation analysis on optimizing weakly discriminative features. 
These features were obtained from the model after the 2nd iteration on the artpainting task of the PACS dataset.
  }
  \label{fig:tsne}
\end{figure}

\textbf{Visual analysis.} 
To provide additional evidence of the impact of incorporating optimization loss on feature distribution in low data resource scenarios, 
we utilized t-SNE to visualize the features on the PACS dataset, as depicted in the Fig. \ref{fig:tsne}. 
A noteworthy observation is that, due to the influence of domain adversarial samples and weakly discriminative feature optimization, there is a greater distinction in inter-class distances among different categories.
This finding demonstrates the effectiveness of our approach in increasing the inter-class distance and mitigating the risk of misclassifying cross-domain samples. 

\section{Limitations}
We argue that the DAAL algorithm exhibits two limitations. 
The first limitation is overfitting due to distribution bias between cross-domain features. 
When there are substantial distribution discrepancies between the target domain and the source domain, our optimization scheme cannot affect the target domain features.
In experiments, as the number of training samples increases, the model's performance in the target domain fails to enhance and may even deteriorate. 
To mitigate this problem, adjusting hyperparameters or augmenting data samples could be necessary.
Additionally, our method of increasing inter-class distances relies on the availability of a sufficient feature space. 
Conversely, when the feature space is limited and there is an abundance of target categories, the enhancement of inter-class distances on the training set remains limited.
This restriction also limits the model's ability to generalize effectively on the test set.
In such scenarios, techniques aiming to decrease intra-class distances would prove advantageous.

\section{Conclusion} 
In this paper, we propose a domain adversarial active learning algorithm for domain generalization classification tasks. 
The algorithm's design aims to leverage the varying contributions of different samples to enhance the model's generalization ability. 
Sepecifically, we introduce two aspects of our work. 
First, for sample selection, we develop a domain adversarial method that use two metric functions to identify challenging samples for domain generalization. 
Second, during the model training process, our focus is on optimizing weakly discriminative feature subsets within each source domain, thereby increasing the inter-class distance within the same domain. 
Finally, we conduct comparative experiments to validate the effectiveness of the proposed method against existing domain generalization and active learning.

\section*{Acknowledgments}
This work is supported by the National Natural Science Foundation of China (No. 72071145).



\bibliographystyle{IEEEtran}
\bibliography{active_learning}
%

\end{document}